\documentclass[twoside,11pt,table,xcdraw]{article}

\usepackage[table,xcdraw]{xcolor}

\usepackage[preprint]{jmlr2e}
\usepackage{xspace}
\usepackage{tabularx}
\usepackage{booktabs}
\usepackage{amsmath}
\usepackage{multirow}
\usepackage{tcolorbox}
\tcbuselibrary{skins}
\usepackage{xparse}

\usepackage{setspace} %
\usepackage{bbm} %

\newcommand{\commenttext}[1]{\small{\bf [ #1 ]}}

\newcommand{\ozan}[1]{\textcolor{orange}{}}
\newcommand{\shijie}[1]{\textcolor{red}{}}
\newcommand{\david}[1]{\textcolor{olive}{}}
\newcommand{\vadim}[1]{\textcolor{blue}{}}
\newcommand{\markd}[1]{\textcolor{purple}{}}
\newcommand{\anju}[1]{\textcolor{magenta}{}}
\newcommand{\anjumust}[1]{\textcolor{red}{}}
\renewcommand{\cite}[1]{\citep{#1}}

\renewcommand{\commenttext}[1]{}

\newcommand{\LN}{\mathop{\mathrm{LN}}\nolimits}
\newcommand{\SA}{\mathop{\mathrm{SA}}\nolimits}
\newcommand{\FFN}{\mathop{\mathrm{FFN}}\nolimits}

\newcommand{\bloom}{BLOOM\xspace}
\newcommand{\gptneox}{GPT-NeoX\xspace}
\newcommand{\opt}[0]{OPT$_{\footnotesize{\textnormal{66B}}}$\xspace}
\newcommand{\palm}[0]{PaLM$_{\footnotesize{\textnormal{540B}}}$}
\newcommand{\bloombig}[0]{BLOOM$_{\footnotesize{\textnormal{176B}}}$}
\newcommand{\evalcohort}[0]{\bloombig{}, \gptneox{}, and \opt{}}

\usepackage{cleveref}
\crefname{section}{\S}{\S\S}
\crefname{table}{Table}{}
\crefname{figure}{Figure}{}
\crefname{algorithm}{Algorithm}{}
\crefname{equation}{Equation}{}
\crefname{appendix}{Appendix}{}
\crefformat{section}{\S#2#1#3}  %

\newcommand{\dataset}{\textsc{FinPile}\xspace}
\newcommand{\model}{\textsc{BloombergGPT}\xspace}
\newcommand{\shortmodel}{\textsc{BloombergGPT}\xspace}

\newcommand{\bhalf}{BF16\xspace}
\newcommand{\half}{FP16\xspace}
\newcommand{\full}{FP32\xspace}

\usepackage{textcomp}
\newcommand{\roughly}{\raisebox{0.5ex}{\texttildelow}}

\makeatletter
\newcommand*\bigcdot{\mathpalette\bigcdot@{.5}}
\newcommand*\bigcdot@[2]{\mathbin{\vcenter{\hbox{\scalebox{#2}{\hspace{0.25em}$\m@th#1\bullet$}}}}}
\makeatother

\NewDocumentCommand\emojileaf{}{
    $\vcenter{\hbox{\includegraphics[height=1em]{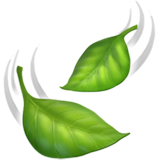}}}$
}
\NewDocumentCommand\emojielbow{}{
    $\vcenter{\hbox{\includegraphics[height=1em]{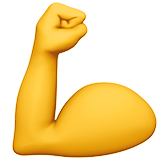}}}$
}
\NewDocumentCommand\emojislide{}{
    $\vcenter{\hbox{\includegraphics[height=1em]{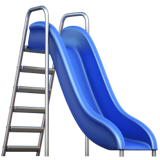}}}$
}
\NewDocumentCommand\emojisuspense{}{
    $\vcenter{\hbox{\includegraphics[height=1em]{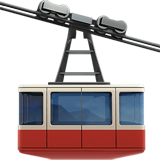}}}$
}

\firstpageno{1}

\begin{document}

\title{\model: A Large Language Model for Finance}

\author{Shijie Wu$^{1,}$\thanks{Co-first authors. Corresponding email: \texttt{{airesearch@bloomberg.net}}} , Ozan \.Irsoy$^{1,*}$, Steven Lu$^{1,*}$, Vadim Dabravolski$^1$, Mark Dredze$^{1,3}$, Sebastian Gehrmann$^1$, Prabhanjan Kambadur$^1$, David Rosenberg$^2$, Gideon Mann$^1$\\ 
        {\small $^1$ Bloomberg, New York, NY USA} \\
        {\small $^2$ Bloomberg, Toronto, ON Canada} \\
        {\small $^3$ Computer Science, Johns Hopkins University, Baltimore, MD USA}
        }

\maketitle

\begin{abstract}%

\noindent The use of NLP in the realm of financial technology is broad and complex, with applications ranging from sentiment analysis and named entity recognition to question answering. Large Language Models (LLMs) have been shown to be effective on a variety of tasks; however, no LLM specialized for the financial domain has been reported in literature.
In this work, we present \model, a 50 billion parameter language model that is trained on a wide range of financial data.
We construct a 363 billion token dataset based on Bloomberg's extensive data sources, perhaps the largest domain-specific dataset yet, augmented with 345 billion tokens from general purpose datasets.
We validate \model on standard LLM benchmarks, open financial benchmarks, and a suite of internal benchmarks that most accurately reflect our intended usage.
Our mixed dataset training leads to a model that outperforms existing models on financial tasks by significant margins without sacrificing performance on general LLM benchmarks.
Additionally, we explain our modeling choices, training process, and evaluation methodology.
We release Training Chronicles (\cref{sec:chron}) detailing our experience in training \model.
\end{abstract}

\tableofcontents 
 
\pagebreak

\section{Introduction}
The release of GPT-3 in 2020 \citep{GPT3} demonstrated the powerful benefits of training very large auto-regressive language models (LLMs). 
GPT-3 had 175 billion parameters, a hundredfold increase over the previous GPT-2 model, and did remarkably well across a wide range of now popular LLM tasks, including reading comprehension, open-ended question answering, and code generation. 
This performance has been replicated across several other models \citep{chowdhery2022palm,bloom,opt-zhang}.
Furthermore, evidence suggests that large models exhibit emergent behaviors; growth allows them to acquire abilities not present in smaller models \citep{emergent-abilities}. A notable example of emergent behavior is the ability to perform tasks via few-shot prompting, where a model can learn a task from just a few examples. This ability improves well-above random as we increase the size of language models. Broadly speaking, few-shot prompting dramatically expands the range of tasks supported by models and lowers the barrier to entry for users seeking automation for new language tasks.

After GPT-3, models grew in size to 280 billion~\citep[Gopher,][]{gopher-rae}, 540 billion~\citep[PaLM,][]{chowdhery2022palm}, and 1 trillion parameters \citep[Megatron,][]{https://doi.org/10.48550/arxiv.2205.05198}. 
Work also explored other important aspects of achieving a high-performing LLM, such as different training objectives \citep{ul2-tay}, multilingual models \citep{bloom}, more efficient and smaller models \citep{black-etal-2022-gpt}, and finding data and parameter-efficient training sizes \citep{hoffmann2022chinchilla}.

These efforts have almost exclusively focused on general LLMs, trained on datasets that cover a broad range of topics and domains. While these have included some datasets for specialized domains (e.g., code \citep{chen2021codex} or biomedical articles \cite{the-pile}) the focus has been on building LLMs with broad capabilities.  Recent efforts training models using only domain-specific data have yielded models that, while much smaller, beat general purpose LLMs on tasks within those domains, such as science \cite{galactica-taylor} and medicine \cite{biomedlm,Luo_2022,lehman-clinicalt5}. These findings motivate further development of models focused on specific domains.

Financial Technology (FinTech) is a large and growing area with NLP technologies having an increasingly important role \cite{xing2018natural,fisher2016natural,dredze2016twitter}. Financial NLP tasks \cite{shah-etal-2022-flue} include sentiment analysis \cite{araci2019finbert}, named entity recognition \cite{salinas-alvarado-etal-2015-domain}, news classification \cite{DBLP:journals/corr/abs-2009-04202}, and question answering \cite{chen-etal-2021-finqa,chen-etal-2022-convfinqa}. While the range of tasks is similar to those found in general NLP benchmarks, the complexity and terminology of the financial domain warrant a domain-specific system. For all of the reasons generative LLMs are attractive in general -- few-shot learning, text generation, conversational systems, etc. -- it would be valuable to have a LLM focused on the financial domain. While there are masked language models tuned for the financial domain~\cite{araci2019finbert}, no LLM has been tuned for or evaluated on tasks for this domain.

\subsection{\model}
We train \model, a 50 billion parameter language model that supports a wide range of tasks within the financial industry. 
Rather than building a general-purpose LLM, or a small LLM exclusively on domain-specific data, we take a mixed approach. 
General models cover many domains, are able to perform at a high level across a wide variety of tasks, and obviate the need for specialization during training time. However, results from existing domain-specific models show that general models cannot replace them. At Bloomberg, we support a very large and diverse set of tasks, well served by a general model, but the vast majority of our applications are within the financial domain, better served by a specific model. 
For that reason, we set out to build a model that achieves best-in-class results on financial benchmarks, while also maintaining competitive performance on general-purpose LLM benchmarks. 

We achieve this goal by constructing the largest domain-specific dataset yet, drawing on existing data creation, collection, and curation resources at Bloomberg. As Bloomberg is primarily a financial data company, our data analysts have collected and curated financial language documents over the span of forty years. We have extensive archives of financial data that cover a range of topics, with careful tracking of data sources and usage rights.
We add this data to public datasets to create a large training corpus with over 700 billion tokens.
Using a portion of this training corpus, we train a BLOOM-style, 50 billion parameter model designed based on guidelines from \citet{hoffmann2022chinchilla} and \citet{le-scao-etal-2022-language}. We validate the model on standard LLM benchmarks, open financial benchmarks, and a suite of Bloomberg-internal benchmarks that most accurately reflect our intended use cases. 
Our results demonstrate that our mixed training approach leads to a model that vastly outperforms existing models on in-domain financial tasks while being on par or better on general NLP benchmarks.

\subsection{Broader Contributions}

Beyond the construction of a LLM for financial data, our goal is to contribute to the broader research community. Specifically, our experience documented in this paper provides evidence that further develops the community's understanding of several open questions in the literature.

\paragraph{Domain-specific LLMs.}
The few existing domain-specific LLMs are trained exclusively on domain-specific data sources \citep{Luo_2022,biomedlm,galactica-taylor}, or adapt a very large general purpose model to domain-specific tasks \citep{medpalm,minerva}. Our alternative approach -- training an LLM on both domain-specific and general data sources -- has not been studied so far. The resulting model does very well on domain-specific tasks, but also maintains strong performance on general-purpose benchmarks.

\paragraph{Training data.}
Nearly all language models rely in large part on web-scraped data, such as C4 \citep{JMLR:v21:20-074} and The Pile \citep{the-pile} (which includes OpenWebText2). This data may be cleaned or subsetted in various ways before use \cite{llama,gopher,bloom,bloom-data-governance}, but issues of data duplication \cite{carlini2021extracting} and toxic language remain \cite{welbl-etal-2021-challenges-detoxifying}. Our training data is unusual for LLM training in that it includes a significant amount of curated and prepared data from reliable sources.

\paragraph{Evaluation.}
LLM evaluation remains a challenging and evolving problem \cite{https://doi.org/10.48550/arxiv.2202.06935,https://doi.org/10.48550/arxiv.2209.12356}, with new benchmarks trying to standardize evaluation across models \citep{HELM,bigbench}. However, for domain-specific tasks, there remains a mismatch between evaluation and actual use cases. Evaluations are built on available datasets and not necessarily on how the model will be used in practice. We provide results on both public financial NLP benchmarks \citep{shah-etal-2022-flue,chen-etal-2021-finqa} as well as a selection of internal Bloomberg tasks, which are better aligned with our intended use cases and directly evaluate our model's ability to perform tasks of interest.

\paragraph{Model Size.}
Early LLMs made a single training pass over a corpus of 200-400 billion tokens \citep{GPT3} and 
\citet{hoffmann2022chinchilla} posited that models were undertrained, instead focusing on training smaller models with more data, a strategy most recently employed by \citet{llama}. We select a model size 
motivated by \citet{hoffmann2022chinchilla} and train a 50 billion parameter model on 569 billion tokens from our corpus of over 700 billion tokens to produce a model that is competitive with larger models.

\paragraph{Tokenizer.}
After assembling training data, the critical step of tokenization transforms the text into a format suitable for the language model. The importance of this step is often overlooked \cite{https://doi.org/10.48550/arxiv.2112.10508}, and many older LLMs use the same tokenizer and vocabulary, meaning that we have little evidence to support other tokenizers. We take a different approach
and use a Unigram model instead of greedy merge-based sub-word tokenizers since it saves probabilities allowing for smarter tokenization at inference time \citep{kudo-2018-subword}.

\paragraph{Model Building Challenges.}
GPT-3 and subsequent models were the work of large teams and required an enormous amount of computation. Initial work to reproduce these results, such as OPT \cite{opt-zhang}, did not match the performance of the original model. With the release of each subsequent model, the community's understanding, experience, and software tools increase. In developing \model, we benefited from existing code developed as part of the BLOOM effort \cite{bloom}, showing that a moderately sized team can produce a competitive model on domain-specific data. 
We describe our experiences training \model in detail to support future training efforts and address each of the above topics.

\section{Dataset}
\label{sec:dataset}
To train \model, we construct ``\dataset'', a comprehensive dataset consisting of a range of English financial documents including news, filings, press releases, web-scraped financial documents, and social media drawn from the Bloomberg archives.
These documents have been acquired through our business process over the past two decades.
We augment \dataset with public data widely used to train LLMs. The result is a training corpus that is roughly half domain-specific text and half general-purpose text. For a breakdown of the full training set, see \cref{tab:data}.
To improve data quality, we de-duplicate each dataset (The Pile, C4, Wikipedia, \dataset) according to \citet{lee-etal-2022-deduplicating}; as a side-effect, the statistics reported in \cref{tab:data} might be different from those reported in other papers.

\begin{table}
\centering
\begin{tabular}{@{}lrrrrrr@{}}\toprule
\multicolumn{1}{c}{\textbf{Dataset}} &
  \multicolumn{1}{c}{\textbf{\begin{tabular}[c]{@{}c@{}}Docs\\ 1e4\end{tabular}}} &
  \multicolumn{1}{c}{\textbf{C/D}} &
  \multicolumn{1}{c}{\textbf{\begin{tabular}[c]{@{}c@{}}Chars\\ 1e8\end{tabular}}} &
  \multicolumn{1}{c}{\textbf{C/T}} &
  \multicolumn{1}{c}{\textbf{\begin{tabular}[c]{@{}c@{}}Toks\\ 1e8\end{tabular}}} &
  \multicolumn{1}{c}{\textbf{T\%}} \\ \midrule
\textit{\dataset }   & \textit{175,886} & \textit{1,017} & \textit{17,883} & \textit{4.92} & \textit{3,635} & \textit{51.27\%}  \\ \cmidrule(r){1-1}
Web                & 158,250          & 933            & 14,768          & 4.96          & 2,978          & 42.01\%           \\
News               & 10,040           & 1,665          & 1,672           & 4.44          & 376            & 5.31\%            \\
Filings            & 3,335            & 2,340          & 780             & 5.39          & 145            & 2.04\%            \\
Press              & 1,265            & 3,443          & 435             & 5.06          & 86             & 1.21\%            \\
Bloomberg          & 2,996            & 758            & 227             & 4.60          & 49             & 0.70\%            \\ \midrule
\textit{PUBLIC}    & \textit{50,744}  & \textit{3,314} & \textit{16,818} & \textit{4.87} & \textit{3,454} & \textit{48.73\%}  \\ \cmidrule(r){1-1}
C4                 & 34,832           & 2,206          & 7,683           & 5.56          & 1,381          & 19.48\%           \\ \cmidrule(r){1-1}
Pile-CC            & 5,255            & 4,401          & 2,312           & 5.42          & 427            & 6.02\%            \\
GitHub             & 1,428            & 5,364          & 766             & 3.38          & 227            & 3.20\%            \\
Books3             & 19               & 552,398        & 1,064           & 4.97          & 214            & 3.02\%            \\
PubMed Central     & 294              & 32,181         & 947             & 4.51          & 210            & 2.96\%            \\
ArXiv              & 124              & 47,819         & 591             & 3.56          & 166            & 2.35\%            \\
OpenWebText2       & 1,684            & 3,850          & 648             & 5.07          & 128            & 1.80\%            \\
FreeLaw            & 349              & 15,381         & 537             & 4.99          & 108            & 1.52\%            \\
StackExchange      & 1,538            & 2,201          & 339             & 4.17          & 81             & 1.15\%            \\
DM Mathematics     & 100              & 8,193          & 82              & 1.92          & 43             & 0.60\%            \\
Wikipedia (en)     & 590              & 2,988          & 176             & 4.65          & 38             & 0.53\%            \\
USPTO Backgrounds  & 517              & 4,339          & 224             & 6.18          & 36             & 0.51\%            \\
PubMed Abstracts   & 1,527            & 1,333          & 204             & 5.77          & 35             & 0.50\%            \\
OpenSubtitles      & 38               & 31,055         & 119             & 4.90          & 24             & 0.34\%            \\
Gutenberg (PG-19)  & 3                & 399,351        & 112             & 4.89          & 23             & 0.32\%            \\
Ubuntu IRC         & 1                & 539,222        & 56              & 3.16          & 18             & 0.25\%            \\
EuroParl           & 7                & 65,053         & 45              & 2.93          & 15             & 0.21\%            \\
YouTubeSubtitles   & 17               & 19,831         & 33              & 2.54          & 13             & 0.19\%            \\
BookCorpus2        & 2                & 370,384        & 65              & 5.36          & 12             & 0.17\%            \\
HackerNews         & 82               & 5,009          & 41              & 4.87          & 8              & 0.12\%            \\
PhilPapers         & 3                & 74,827         & 23              & 4.21          & 6              & 0.08\%            \\
NIH ExPorter       & 92               & 2,165          & 20              & 6.65          & 3              & 0.04\%            \\
Enron Emails       & 24               & 1,882          & 5               & 3.90          & 1              & 0.02\%            \\ \cmidrule(r){1-1}
Wikipedia (7/1/22) & 2,218            & 3,271          & 726             & 3.06          & 237            & 3.35\%            \\ \midrule
\textit{TOTAL}     & \textit{226,631} & \textit{1,531} & \textit{34,701} & \textit{4.89} & \textit{7,089} & \textit{100.00\%}\\\bottomrule
\end{tabular}     
\caption{Breakdown of the full training set used to train \model. The statistics provided are the average number of characters per document (``C/D''),
the average number of characters per token (``C/T''),
 and the percentage of the overall tokens (``T\%''). Units for each column are denoted in the header.}
\label{tab:data}
\end{table}

\subsection{Financial Datasets (363B tokens -- 51.27\% of training)}
The Bloomberg Terminal has provided access to a comprehensive set of diverse structured and unstructured financial data and analytics for the past four decades.
In serving this mission, Bloomberg analysts have curated a set of financial documents that were either created internally or acquired from external sources.
We utilize this extensive collection of curated and maintained documents to create \dataset, which consists of company filings, financial news, and other data relevant to the financial markets.

Some documents included in the \dataset, such as company filings, are available to the general public, although collecting these documents and pre-processing them for LLM training is a non-trivial task.
Other documents, such as (a subset of) Bloomberg news, must be purchased.
The rest of the documents are private and available, among other sources, through the Bloomberg Terminal.
Finally, we clean this data to strip off markup, special formatting, and templates.

Note that each document in \dataset is time-stamped, with dates ranging from 2007-03-01 to 2022-07-31; the quality and quantity of documents increase over this time range.
 While we do not utilize date information in this work, we plan to use it in the future, such as for evaluation of what the model learns about different time periods.
While we cannot release \dataset, our experience training on a large, carefully curated, and clean domain-specific dataset may provide helpful insights to the community on the advantages and challenges of building a financial LLM in particular, and a domain-specific model in general.
We provide a breakdown and analysis of \dataset in \cref{tab:bbpile} and a brief description of the types of data included below.

\begin{table}[t]
\centering
\begin{tabular}{@{}lrrrrrr@{}}
\toprule
Date           & Bloomberg  & Filings & News  & Press & Web    & \textit{Total}  \\ \midrule
2007 {[}03-{]} & 276 & 73      & 892   & 523   & 2,667  & \textit{4,431}  \\
2008           & 351 & 91      & 1,621 & 628   & 9,003  & \textit{11,695} \\
2009           & 293 & 93      & 1,791 & 528   & 9,179  & \textit{11,883} \\
2010           & 292 & 111     & 1,917 & 527   & 11,388 & \textit{14,236} \\
2011           & 335 & 117     & 2,264 & 548   & 13,643 & \textit{16,907} \\
2012           & 403 & 105     & 2,502 & 529   & 15,015 & \textit{18,554} \\
2013           & 415 & 87      & 2,437 & 441   & 17,230 & \textit{20,610} \\
2014           & 396 & 251     & 2,458 & 437   & 18,510 & \textit{22,052} \\
2015           & 358 & 1,639   & 2,371 & 427   & 20,782 & \textit{25,576} \\
2016           & 324 & 1,891   & 2,509 & 418   & 24,337 & \textit{29,478} \\
2017           & 294 & 2,294   & 2,567 & 398   & 25,283 & \textit{30,837} \\
2018           & 275 & 1,791   & 2,702 & 420   & 26,027 & \textit{31,214} \\
2019           & 263 & 1,662   & 3,102 & 504   & 27,195 & \textit{32,726} \\
2020           & 277 & 1,632   & 2,794 & 805   & 30,928 & \textit{36,435} \\
2021           & 247 & 1,767   & 3,515 & 938   & 29,749 & \textit{36,215} \\
2022 {[}-07{]} & 140            & 882             & 2,206           & 531            & 16,872           & \textit{20,631}  \\ \midrule
\textbf{}      & \textit{4,939} & \textit{14,486} & \textit{37,647} & \textit{8,602} & \textit{297,807} & \textit{363,482} \\ \bottomrule
\end{tabular}
\caption{The number of tokens (in millions) contained within documents in \dataset, organized by year (rows) and type (column). Units are millions of tokens.
}
\label{tab:bbpile}
\end{table}

\subsubsection{Web (298B tokens -- 42.01\% of training)}

Bloomberg collects web content by identifying sites that contain financially relevant information.
While this category makes up the majority of \dataset, its classifications are rough, with content classified mainly by the location of the web domain.
Within these location-specific sources, e.g. ``US'' (15.95\% of total), ``Asia-Pac'' (4.72\% of total), and ``UK'' (1.98\% of total), document types are highly varied as would be expected from a web crawl.
While web sources are common in existing public LLM training datasets, Bloomberg's web crawl is focused on high-quality websites that have financially relevant information, as opposed to a general-purpose crawl of the web.

\subsubsection{News (38B tokens -- 5.31\% of training)}
The News category includes all news sources excluding news articles written by Bloomberg journalists.
Overall, there are hundreds of English news sources in \dataset including ``Bloomberg Transcripts'' (0.41\% of total), which are transcripts of Bloomberg TV news.
Generally, the content in this dataset comes from reputable sources of news that are relevant to the financial community so as to maintain factuality and reduce bias.

\subsubsection{Filings (14B tokens -- 2.04\% of training)}
Company Filings are financial statements prepared by (public) companies and made available to the general public. %
In some countries, like the US, public companies are mandated to prepare and submit their financial statements on a regular cadence; e.g., 10-K annual reports and 10-Q quarterly reports.
In our dataset, a majority of the filings come from EDGAR, which is the SEC's online database (1.90\% of total).
Filings are typically long PDF documents with tables and charts that are dense in financial information, which are processed and normalized in Bloomberg.
Filings are substantially different from the types of documents typically used to train LLMs, but contain critically important information for financial decision-making.

\subsubsection{Press (9B tokens -- 1.21\% of training)}
The Press category contains press releases typically issued by companies that are financially relevant.
Taken together with filings, press releases represent most of the public communications of a company. 
However, unlike filings, press releases are similar to news stories in terms of content and style.

\subsubsection{Bloomberg (5B tokens -- 0.70\% of training)}
This category comprises Bloomberg authored news and other documents such as opinions and analyses.
The largest sources are ``Bloomberg News'' (0.44\% of total) and ``Bloomberg First Word'' (0.13\% of total), the Bloomberg-authored wire of real-time news.
While Bloomberg News covers a wide range of topics, it typically focuses on content relevant to the financial community.
This dataset contains documents of varying lengths.

\subsection{Public Datasets (345B tokens -- 48.73\% of training)}

We use three widely known and available public datasets in our training corpus.

\subsubsection{The Pile (184B tokens -- 25.9\% of training)}
The Pile~\citep{the-pile} is the dataset used in GPT-Neo~\citep{gpt-neo}, GPT-J~\citep{gpt-j}, and \gptneox (20B)~\citep{black-etal-2022-gpt}.
We include The Pile in our training data for the following reasons.
First, it has been used to successfully train an LLM.
Second, it has undergone significant data cleaning and pre-processing.
Third, it includes multiple domains and we believe such diverse data will aid generalization to new domains and may even support training on financial data.
For example, domains such as FreeLaw and GitHub are useful to teams at Bloomberg that work on legal documents and software development, respectively.
Creators of The Pile have deliberately chosen to include duplicate content, with the duplication factor being proportional to the perceived quality of the content.
However, as we deduplicate each of our datasets, the size of The Pile is significantly reduced.
Additionally, note that our tokenizer (\S\ref{sec:tokenization}) is trained on The Pile.

\subsubsection{C4 (138B tokens -- 19.48\% of training)}
The Colossal Clean Crawled Corpus (C4) is a common dataset used to train LLMs, and was introduced to support training T5 \citep{JMLR:v21:20-074}.
Although it overlaps with Pile-CC, C4 is cleaned and processed differently; hence, we feel that including C4 in addition to The Pile can add value more than duplicated documents would.
We find that C4 contains high-quality natural language documents due to the layers of cleaning, though others have noted that the distribution across web domains is unusual, with a high fraction of data stemming from patents \cite{dodge-etal-2021-documenting}.

\subsubsection{Wikipedia (24B tokens -- 3.35\% of training)}
Both The Pile and C4 include out-of-date copies of Wikipedia, so it could be beneficial for the factuality of the model to have up-to-date Wikipedia pages included.
Therefore, we include a dump of English Wikipedia from July 1, 2022. 
This dataset is tokenized quite inefficiently (3.06 characters per token), indicating an above-average amount of markup, which suggests that further cleaning might benefit future model training.

\subsection{Tokenization}
\label{sec:tokenization}
We choose the Unigram tokenizer \citep{kudo-2018-subword} instead of a greedy merge-based
sub-word tokenizer, such as Byte Pair Encoding (BPE) \citep{sennrich-etal-2016-neural} or
Wordpiece \citep{schuster2012japanese,wu2016google}, based on promising results in \citet{kudo-richardson-2018-sentencepiece} and \citet{bostrom-durrett-2020-byte}.
Following GPT-2 \cite{gpt-2-radford}, we treat our data as a sequence of bytes rather than
Unicode characters, and we include each of the 256 bytes as tokens.
In a pretokenization step, the input byte sequence is broken into chunks by greedily
matching the following regular expression: \texttt{[ A-Za-z]+|[0-9]|[\textasciicircum
  A-Za-z0-9]+}. This follows GPT-2 in preventing multiple
character classes from appearing in a single
token. However, we include spaces in the alphabetic chunks, which allows multi-word tokens to be learned, increasing information density and reducing context lengths. The pretokenization follows the approach of PaLM \cite{chowdhery2022palm} in placing
each digit in its own chunk, with the hope that this will lead to better
handling of numbers.
We train our tokenizer on The Pile \cite{the-pile} as it draws from diverse domains, including code and academic papers, in proportions that suit our use case.

\paragraph{Parallel Tokenizer Training.}
The Unigram tokenizer implementation is too inefficient to process the entire Pile dataset at once, so we use a split
and merge approach. We split each of the 22 domains in the Pile into 256 chunks
of roughly equal size. We then train a Unigram tokenizer with a vocabulary
size of 65,536 ($2^{16}$) on each of the $22 \times 256$ (total $=5,632$) chunks. We
hierarchically merge the individual tokenizers by first merging the 256
tokenizers from each domain, and then combining the 22 resulting tokenizers to
get the final tokenizer.

Unigram tokenizers amount to probability distributions
over tokens (i.e. unigram language models), and we merge tokenizers
by taking a weighted average of the probabilities of corresponding tokens, with
the weights determined by the relative sizes (in bytes) of the data used to
train the tokenizers. The result is a tokenizer with 7 million tokens. To reduce
the size of the vocabulary to $2^{17}$ tokens, we drop the tokens with
the smallest probabilities and renormalize. To ensure we do not need an
out-of-vocabulary token, we also add as tokens the 36 (of 256 possible) bytes that do not
occur in The Pile, along with an \texttt{<|endoftext|>} token.

There are various considerations in choosing the vocabulary size. One advantage of a large vocabulary for LLMs is that more information can fit into the context window. On the other hand, there is overhead with a larger vocabulary: a larger proportion of model parameters are required for token embedding. We select our vocabulary size of $2^{17}$ tokens based on experiments with vocabulary ranging from 25,000 to 550,000. For each vocabulary size, we tokenize the C4 dataset and compute the total size (in bytes) for the dataset, where each token is represented using $\log_2(\text{vocabulary size})$ bits. Our heuristic is to choose the vocabulary size that leads to the smallest encoded representation of C4. This gives us a vocabulary size of 125,000, which we then round up to the nearest power of 2 ($2^{17}$, or 131,072 tokens). Our tokenizer is large, relative to the standard vocabulary size of approximately 50,000 tokens.
For an analysis of tokenization efficiency, see Table \ref{tab:tok}.

\begin{table}[t]
\centering
\begin{tabular}{@{}lrrrrrrr@{}}
\toprule
 &
  \multicolumn{1}{l}{\textbf{BLOOM}} &
  \multicolumn{1}{l|}{\textbf{/ours}} &
  \multicolumn{1}{l}{\textbf{NeoX}} &
  \multicolumn{1}{l|}{\textbf{/ours}} &
  \multicolumn{1}{l}{\textbf{OPT}} &
  \multicolumn{1}{l|}{\textbf{/ours}} &
  \multicolumn{1}{l}{\textbf{\shortmodel}} \\ \midrule
FinPile (old)   & 451          & 110\%          & 460          & 112\%          & 456          & 111\%          & 412          \\
C4             & 166          & 121\%          & 170          & 123\%          & 170          & 123\%          & 138          \\
The Pile       & 203          & 110\%          & 214          & 116\%          & 239          & 130\%          & 184          \\
Wikipedia      & 21           & 88\%           & 23           & 99\%           & 24           & 103\%          & 24           \\
\textit{Total} & \textit{390} & \textit{113\%} & \textit{408} & \textit{118\%} & \textit{434} & \textit{126\%} & \textit{345} \\ \bottomrule
\end{tabular}
\caption{Number of tokens in each training dataset with BLOOM, NeoX, OPT (GPT2), and \model tokenizers. All token counts are in billions (B). Note that an older version of FinPile was used for this count, so token numbers will not match earlier tables.}
\label{tab:tok}
\end{table}

\section{Model}

\subsection{Architecture}

Our model is a decoder-only causal language model based on BLOOM \citep{bloom}. We present an overview of the architecture, with full details in~\cref{sec:architecture}.

The model contains 70 layers of transformer decoder blocks defined as follows:
\begin{align*}
\bar{h}_\ell &= h_{\ell-1} + \SA(\LN( h_{\ell-1} )) \\
h_\ell &= \bar{h}_\ell + \FFN(\LN( \bar{h}_\ell ))
\end{align*}
where $\SA$ is multi-head self-attention, $\LN$ is layer-normalization, and $\FFN$ is a feed-forward network with 1-hidden layer. Inside FFN, the non-linear function is GELU \citep{hendrycks2016gaussian}. ALiBi positional encoding is applied through additive biases at the self-attention component of the transformer network \citep{le-scao-etal-2022-language}. The input token embeddings are tied to the linear mapping before the final softmax. Following \citet{le-scao-etal-2022-language} and first used in \citet{dettmers8}, the model has an additional layer normalization after token embeddings, formally:
\begin{align*}
\bar{h}_1 &= \LN^{em}(h_0) + \SA(\LN( \LN^{em} ( h_0 ) )),
\end{align*}
where $h_0$ is the initial token embedding and $\LN^{em}$ is the new component of \emph{em}bedding layer-normalization. Notice that the second term includes two consecutive layer-normalizations. 

\begin{table}
\centering
\begin{tabular}{lr} \toprule
\textbf{Shape} & \\ \midrule 
Number of Layers &  70 \\
Number of Heads & 40 \\
Vocabulary Size & 131,072 \\
Hidden Dimension & 7,680 \\
Total Parameters & 50.6B \\ \midrule
\textbf{Hyperparameters} & \\ \midrule
Max Learning Rate & 6e-5 \\ 
Final Learning Rate & 6e-6 \\ 
Learning Rate schedule & cosine decay \\ 
Gradient Clipping & 0.3 \\ \midrule
\textbf{Training} & \\ \midrule
Tokens & 569B \\
Hardware & $64 \times 8$ A100 40GB \\ 
Throughput & 32.5 sec/step \\
avg. TFLOPs & 102 \\ 
total FLOPS & ~2.36e23 \\ 
\bottomrule
\end{tabular}
\caption{A summary of the hyper-parameters and their values for \model.}
\label{tab:parameter_summary}
\end{table}

\subsection{Model Scaling}

\begin{figure}[t]
    \centering
    \includegraphics[width=\textwidth]{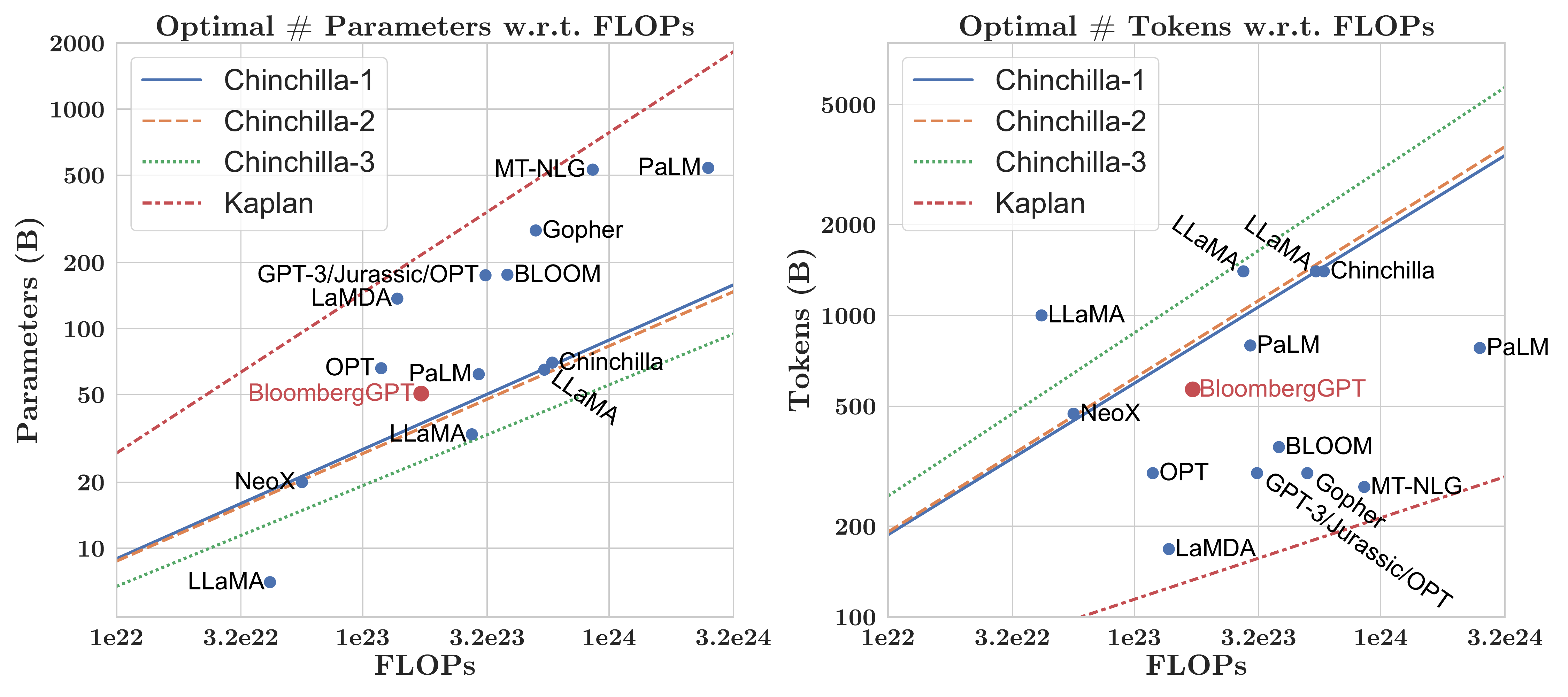}
    \caption{\citet{scaling-kaplan} and Chinchilla scaling laws with prior large language model and \model parameter and data sizes. We adopt the style from \citet{hoffmann2022chinchilla}.}
    \label{fig:scaling}
\end{figure}

\paragraph{Size.}
The size of our model is based on Chinchilla scaling laws ~\citep{hoffmann2022chinchilla}, in particular their Approach 1 and Approach 2. We start with a total compute budget of 1.3M GPU hours on 40GB A100 GPUs. Since we adopt activation checkpointing to reduce our memory footprint, this costs us an additional 0.33x TFLOPs per iteration due to repeated forward passes. To account for this additional cost, we plug in 0.75 $\times$ 1.3M into Chinchilla equations instead of the full amount.

From \citet{hoffmann2022chinchilla}, we use the data reported in Table~3 for Approach 1 and Table~A3 for Approach 2, and fit regression lines to their log-scaled versions. This gives us:
\begin{align*}
    & \text{Approach 1} & Parameters &= \exp_{10}( \log_{10}(FLOPs) \cdot 0.498 - 1.004 ) = 52.993\text{B}\\
    &                   & Tokens     &= \exp_{10}( \log_{10}(FLOPs) \cdot 0.502 + 0.229 ) = 1111.112\text{B}\\
    & \text{Approach 2} & Parameters &= \exp_{10}( \log_{10}(FLOPs) \cdot 0.490 - 0.839 ) = 49.753\text{B}\\
    &                   & Tokens     &= \exp_{10}( \log_{10}(FLOPs) \cdot 0.510 + 0.062 ) = 1175.766\text{B}
\end{align*}
These calculations imply that our dataset of \roughly 700B tokens is too small for a ``Chinchilla optimal'' configuration given our compute budget (assuming just one pass through the data).\footnote{The scaling law derived by Chinchilla is tokenizer-specific. Our tokenizer can encode the same document more compactly due to the support of multi-word expressions and the larger vocabulary size. It's still an open question how well these scaling laws transfer across tokenizers, and how vocabulary size impacts token and parameter trade-offs assuming fixed compute. We leave this exploration to future work.
}
While we can increase the amount of general-purpose training data, we are limited in the amount of domain-specific training data at our disposal. \dataset is already among the largest domain-specific training sets, and we do not want it to represent less than half of our total training. 
 
 Since we are data limited, we choose the largest model that we can, while ensuring that we can train on all our tokens and still leave \roughly 30\% of the total compute budget as a buffer for unforeseen failures, retries, and restarts. This leads us to a 50B parameter model, which is also roughly the Chinchilla optimal size for our compute budget. \cref{fig:scaling} provides a summary of the scaling laws and how \model compares to other models.

\paragraph{Shape.} To determine how to allocate the 50B parameters to different model components (i.e., the ``shape'' of our model), we follow \citet{levine2020limits}, who propose that for a total number of self-attention layers $L$, the optimal hidden dimension $D$ is obtained by:
\begin{align*}
    D &= \exp(5.039) \exp(0.0555 \cdot L)
\end{align*}
We sweep $L$ over a range of integer values and pick the $(L, D)$ combination that yields a total of \roughly 50B parameters. This leads to the choice of $L=70$ and $D=7510$ as our target shape parameters. However, we also want to follow the tradition that the hidden dimension is evenly divisible by the number of attention heads, with the quotient giving the attention head dimension.  Furthermore, we want the dimensions to be multiples of 8 to achieve higher performance in Tensor Core operations~\cite{nvidia-mixed-precision}. We settle on 40 heads, each having a dimension of 192, resulting in a total hidden dimension of $D=7680$ and a total of 50.6B parameters. \cref{tab:parameter_summary} provides a summary of the hyper-parameters used in \model.

\subsection{Training Configuration}

\paragraph{Training.} \model is a PyTorch model trained with a standard left-to-right causal language modeling objective. 
Following \citet{GPT3}, we want all our training sequences to be exactly the same length, in our case 2,048 tokens, to maximize GPU utilization. To achieve this, we concatenate all our tokenized training documents with an \texttt{<|endoftext|>} token as a document separator. We then break this token sequence into chunks of 2,048 tokens. Note that with this approach, each training sequence may contain multiple documents from different domains. Also note that, because we're using ALiBi positional encoding, \model can be applied to sequences longer than 2,048 at inference time. For optimization efficiency, training sequences are grouped together into batches, as described in more detail below. 

\paragraph{Optimization.} We use the AdamW optimizer \citep{loshchilov2018decoupled}. We set $\beta_1$ to 0.9, $\beta_2$ to 0.95, and weight decay to 0.1. Following \citet{GPT3}, we set the maximum learning rate to 6e-5 and use the cosine decay learning rate scheduler with linear warmup. We warm up the learning rate in the first 1800 steps. Following \citet{hoffmann2022chinchilla}, the final learning rate is 0.1x the max learning rate, i.e. 6e-6. We also employ batch size warmup \citep{GPT3}: in the first 7,200 steps, we use a batch size of 1,024 (2.1M tokens), then switch to a batch size of 2,048 (4.2M tokens) for the remainder of training. %

We set dropout to 0.0 in all layers in our initial run, although we add dropout later as explained in \S\ref{sec:run}.
The model parameters are randomly initialized to samples from a normal distribution with zero mean and standard deviation $\sqrt{1/(3D)} = 0.006588$ \citep{https://doi.org/10.48550/arxiv.2201.11990}. Following Megatron-LM \citep{shoeybi2019megatron}, we rescale the standard deviation of the second layer in the MLP and the output layer of the attention by $1/ \sqrt{2L}$. We use the technique of \texttt{query\_key\_layer\_scaling} \citep{shoeybi2019megatron}, which was proposed to improve numerical stability for \half  mixed-precision training but may also help in \bhalf. %

\paragraph{Training Instability.}
LLMs optimization requires running convex optimization algorithms over incredibly complex non-convex loss surfaces. Previous work has reported various instabilities while training LLMs. For example, \citet{chowdhery2022palm} found that the loss spiked roughly 20 times while training PaLM, despite the fact that gradient clipping was enabled. They mitigated these issues by re-starting training from a checkpoint roughly 100 steps before the spike started, and then skip 200–500 data batches. They hypothesized that spikes occur due to the combination of specific data batches with a particular model parameter state. 
Similarly, during OPT training, \citet{opt-zhang} noticed spikes in the gradient and activation norms, or divergences in the training perplexity. After these behaviors, they lowered their learning rate, which stabilized these norms and allowed training to continue.
Interestingly, \citet{bloom} report only a single loss spike, from which the model recovered on its own.

\paragraph{Hardware Stack.}
We use the Amazon SageMaker service provided by AWS to train and evaluate \model. We use the latest version available at the time of training and train on a total of 64 p4d.24xlarge instances. Each p4d.24xlarge instance has 8 NVIDIA 40GB A100 GPUs with NVIDIA NVSwitch intra-node connections (600 GB/s) and NVIDIA GPUDirect using AWS Elastic Fabric Adapter (EFA) inter-node connections (400 Gb/s). This yields a total of 512 40GB A100 GPUs.
For quick data access, we use Amazon FSX for Lustre, which supports 
up to 1000 MB/s read and write throughput per TiB storage unit.\looseness=-1

\subsection{Large-scale Optimization}

To train \model, which has a larger memory footprint than available GPU memory on cloud instances, we rely on stage 3 of ZeRO optimization \citep{rajbhandari2020zero}. We utilize the proprietary SageMaker Model Parallelism (SMP) library from AWS, which enables the automatic distribution of large models across multiple GPU devices and instances~\citep{karakus2021amazon}. 
After experimenting with various techniques, we achieve 102 TFLOPs on average and each training step takes 32.5 seconds. We find the following setup to be the best performing in our training.

\paragraph{ZeRO Optimization (stage 3).} ZeRO shards the training state (model parameters, gradients, and optimizer state) across a group of GPUs. We shard a model across 128 GPUs, and we have 4 copies of the model during training.

\paragraph{MiCS.} \citet{misc-sagemaker} decrease training communication overhead and memory requirements for cloud training clusters. MiCS includes such features as hierarchical communication, 2-hop gradient update, scale-aware model partitioning.

\paragraph{Activation Checkpointing.} \citet{chen2016training} minimizes training memory consumption by removing activations at the expense of additional computation during backward passes.
When a layer has activation checkpointing enabled, only the layer input and outputs are kept in memory following a forward pass, while any intermediate tensors are discarded from memory. 
During the backward pass, these intermediate tensors may be recomputed. 
We apply activation checkpointing to each transformer layer.

\paragraph{Mixed Precision Training.} 
To reduce the memory requirements, forward and backward passes are done in \bhalf, while parameters are stored and updated in full precision (\full). The ALiBi matrices are computed in full precision and stored in \bhalf. We also use \full to calculate fused softmax in the Attention block and store its results in \bhalf. Finally, the softmax calculations in the loss function are computed in \full.

\paragraph{Fused Kernels.} Another possibility for optimization is combining composition of several operations into a single GPU operation. This can both reduce peak memory usage by avoiding storage of intermediate results in the computation graph, as well as help improve speed. Similar to Megatron-LM \cite{shoeybi2019megatron}, we use a masked-causal-softmax fused kernel in SMP in the self-attention module.  In practice, we observe 4-5 TFLOPs improvement for speed, and avoid out-of-memory errors given the rest of the configuration.

\section{Training Run}
\label{sec:run}

\begin{figure}[t]
    \centering
    \includegraphics[width=\textwidth]{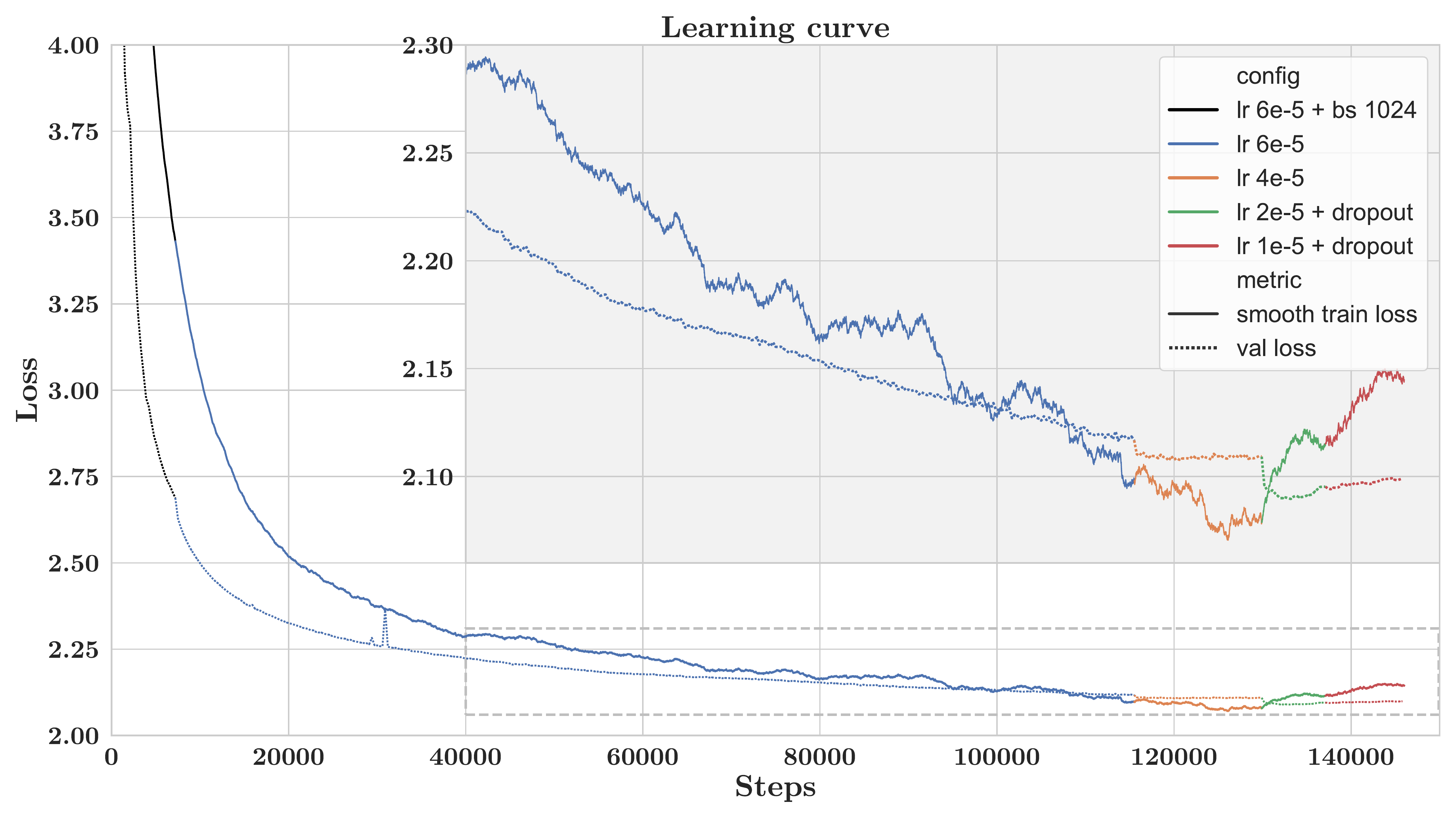}
    \caption{(Smoothed) training and validation losses for \model. Inner plot is a zoomed-in version of the area within dashed rectangle in the outer plot (with shared x-axis). Colors denote different hyperparameter configurations. Styles denote training vs validation loss.}
    \label{fig:training_run}
\end{figure}
The process of training \model involved decisions along the way based on the progress of model training. We share some highlights of this process. A detailed presentation appears in the Training Chronicles (\cref{sec:chron}). 
\cref{fig:training_run} shows the learning curves for both training and validation sets. The solid lines show (smoothed) training loss and the dotted lines show loss on the held-out validation set. Changes in the color of the lines indicate changes to the optimization hyperparameter configurations, either as scheduled, or in response to increasing or stagnating validation loss. This plot shows the path taken by the successful model training run. To present a clear plot, the Figure does not show other attempts with different model configurations, overwritten partial runs after a rollback, or other training strategies not utilized in the final model.

We measured training loss every five steps on the current batch. The raw values vary wildly, causing large jitter when plotted. The plot smoothes the training loss by showing a running average $y_t = \frac{\sum_{i=0}^t x_i \cdot (1-\alpha)^{(t-i)}}{\sum_{i=0}^t (1-\alpha)^{(t-i)}}$ where $\alpha=0.001$. Smoothing is not needed for the validation loss since it is measured on the entire validation set every 300 steps.

We trained for a total of 139,200 steps ($\roughly 53$ days) and ended model training after completing $\roughly 80\%$ of one epoch through our training data (569B tokens out of the 709B tokens available). We ended training early because the loss on our held-out development set was no longer improving, although it's possible that substantially longer training may have yielded further improvements.

We began the run with a warm-up batch size of 1,024 for 7,200 steps, after which we switched to the regular batch size of 2,048 (color changes from black to blue). Change in batch size manifests as a visible curvature change in the validation loss at step 7,200. Most of the remainder of the training performed stably with decreasing training and validation losses. Intervention was required at later stages, after step 115,500, when we observed flat or increasing validation loss. We then applied the following corrective modifications in sequence:
\begin{itemize}
    \setlength\itemsep{-1em}
    \item Step 115,500 (blue to orange): Shrink learning rate to two-thirds \\
    \item Step 129,900 (orange to green): Halve learning rate, and add dropout (with 0.1 probability)\\
    \item Step 137,100 (green to red): Halve learning rate again
\end{itemize}
We ended the run at step 146,000 based on the lack of observable progress on the validation loss. We selected the checkpoint at step 139,200 as the final model based on validation loss and downstream evaluations.

\begin{table}[t]
\centering
\small
\begin{tabular}{@{}lll@{}}
\toprule
Suite & Tasks & What does it measure? \\ \midrule
Public Financial Tasks & 5 & Public datasets in the financial domain \\
Bloomberg Financial Tasks & 12 & NER and sentiment analysis tasks\\
\midrule
Big-bench Hard~\citep{BBH} & 23 & Reasoning and general NLP tasks \\
Knowledge Assessments & 5 & Testing closed-book information recall\\
Reading Comprehension & 5 & Testing open-book tasks \\
Linguistic Tasks & 9 & Not directly user-facing NLP tasks\\
\bottomrule
\end{tabular}
\caption{Evaluation Benchmarks. We evaluate \model on a high-coverage set of standard benchmarks that assess downstream performance, taken from HELM, SuperGLUE, MMLU, and the GPT-3 suite. Since these have significant overlap and/or include each other, we restructure them into the categories presented here. We only evaluate on one setup per dataset. We further assess \model on a suite of internal and public financial tasks.}
\label{tab:eval-sets}
\end{table}

\section{Evaluation}
\label{sec:eval}

We evaluated the performance of \model on two broad categories of tasks: finance-specific and general purpose.
The finance-specific tasks help us test our hypothesis that training on high-quality finance-specific data will yield better results on financial tasks.
The general purpose tasks investigate whether the performance of our model is directly comparable to previously published results.
For financial tasks, we assembled publicly available financial datasets that include a range of NLP tasks.
Then, to directly test \model's ability on Bloomberg tasks of interest, we also included tasks drawn from Bloomberg-internal high-quality evaluation sets for sentiment analysis and named entity recognition. 
For general-purpose tasks, we draw from multiple existing benchmarks and group results into the following categories: BIG-bench Hard, Knowledge Assessments, Reading Comprehension, and Linguistic Tasks. The number of tasks per type and the definitions of the groups are presented in \cref{tab:eval-sets}.

\begin{table}[t]
\small
\centering
\begin{tabular}{@{}l cccc@{}}
\toprule
Name & \# Tokens (B) & \# Params. (B) & Compute \\ \midrule
\model & 569 & 50.6 & 1.00$\times$  \\
\gptneox & 472 & 20 & 0.33$\times$ \\
OPT & 300 & 66 & 0.69$\times$ \\
BLOOM & 366 & 176 & 2.24$\times$ \\ \midrule
GPT-3 & 300 & 175 & 1.82$\times$ \\ 
\bottomrule
\end{tabular}
\caption{Evaluation model cohort. OPT and BLOOM each have multiple sizes available and we report those we evaluated. 
We note that compute numbers are only partially comparable between models: For example, BLOOMs training data is only 1/3 English, and OPT repeated some of its training data. We report GPT-3 results whenever available but did not run it ourselves due to lack of availability.}
\label{tab:eval-cohort}
\end{table}

We compare \model to the three closest models described in \cref{sec:related_works} based on model size, type of training data, overall performance, and most importantly, access. An overview of the model sizes and compute is provided in Table~\ref{tab:eval-cohort}.
\begin{enumerate}
\item \gptneox~\citep{black-etal-2022-gpt}: According to \citet{HELM}, this model is the best performing available model under 50B parameters.
\item \opt~
 \citep{opt-zhang}: We chose to compare to \opt since our model size and structure roughly match, though our model is smaller. 
 \item \bloombig{} \citep{bloom}: While this model is substantially larger than \model, we use the same model architecture and software stack. We note that \bloombig{} is multilingual, so while it is much larger, it also is trained on data from more languages.
\end{enumerate}

\noindent All three models use some of the same general-purpose datasets we use in our training corpus. 
We additionally report results from the original GPT-3~\citep{GPT3} whenever externally available.\footnote{Another related general-purpose model at a comparable size ~\citep[LLaMA,][]{llama}, was released during the preparation of this manuscript, but third-party evaluation results were not available and we haven't received access to the model weights.}

We prefer running models ourselves to ensure identical evaluation setups, and we place any results that have been reported elsewhere and were not run by us into a separated group.
To fairly compare the models, we avoid any tuning of prompts and other techniques that could lead to improved results for some, but not all, models.
For that reason, every task is tested via ``standard'' prompting (shown in \cref{tab:eval-method}), i.e., without any parameter changes to the underlying model, without task descriptions, and without Chain-of-Thought prompting~\citep{wei2022chain}.
The number of few-shot examples presented to the model depends on the task, and we include these details in the respective sections. For each group of results, we further present a win rate similar to \citet{HELM} that represents the fraction of ``wins'' in side-by-side comparisons over individual tasks between all model pairs for which we have run the evaluation ourselves. 

\subsection{Few-shot Methodology}
\label{sec:method}
For tasks where a set of candidates are given, we perform likelihood-based classification, following \citet{GPT3}. We consider three methods for classification: regular, calibration, and normalization. Formally,
\begin{itemize}
    \item Regular: $\arg\max_\alpha p(\alpha|\mathbf{s})$
    \item Calibration: $\arg\max_\alpha p(\alpha|\mathbf{s}) / p(\alpha|\text{``Answer:''})$
    \item Normalization: $\arg\max_\alpha p(\alpha|\mathbf{s}) / \textrm{len}(\alpha)$
\end{itemize}
where $\alpha$ is a candidate, $\mathbf{s}$ is the context, and len measures the number of sub-word tokens. We report the performance of the best method for each model and task. For other tasks, we perform generation via greedy decoding. 

We use the official split and report performance on the test set whenever possible. If the test labels are not publicly available, we report performance on the dev set instead. If an official split for a dataset does not exist, we create train and test splits by selecting 20\% of examples to be the test and the rest as train.
All few-shot context examples are sampled from the training set.
To reduce the variance of few-shot evaluation, we sample different shots for each test example, unless otherwise specified. For the sake of consistency, for each test example, all models have identical surface form as input in our evaluation.

\subsection{Heldout Loss}

\begin{figure}[t]
    \centering
    \includegraphics[width=0.8\textwidth]{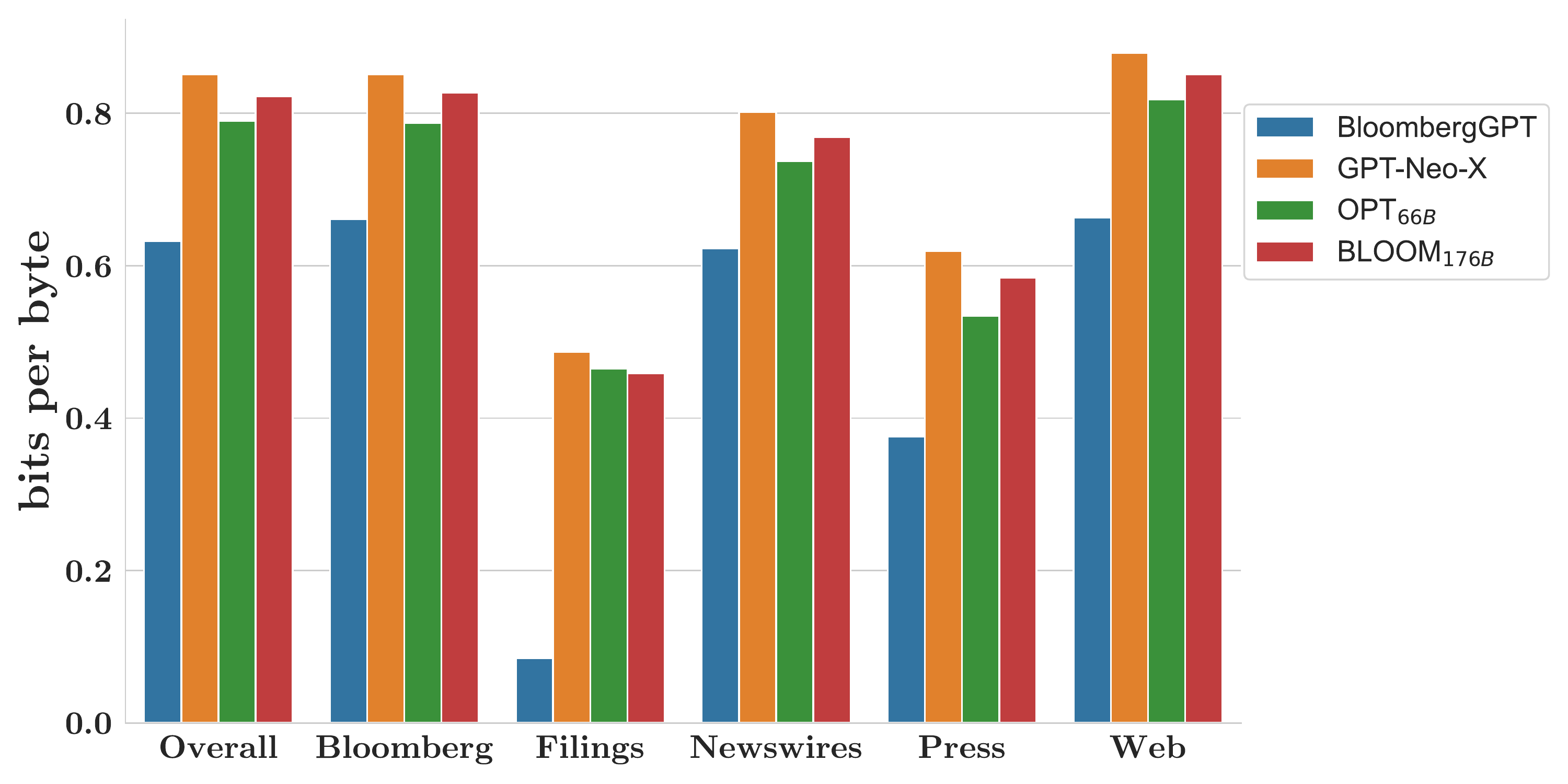}
    \caption{Bits per byte on a heldout test set of each data type in our \dataset (lower is better). The set of documents is held out in time and deduplicated with the training set, such that all of it is completely unseen by \model{}. Regardless, we observe a large gap between the models. The improvement is largest for specialized in-domain documents like Filings.}
    \label{fig:dev_loss}
\end{figure}
We begin by testing how well \model models the language distribution of the in-distribution finance data.
We evaluate the bits per byte of the different models on a heldout dataset that contains examples from all sections of \dataset (described in \S\ref{sec:dataset}).
To limit data leakage and better simulate real-world usage of LLMs, we select a temporally heldout dataset that is strictly further in the future than the training set, and perform deduplication between the training and heldout set.
During evaluation, for documents that are longer than 2,048 tokens, we use a sliding window approach with half window size as context. That means that any token beyond the first 2,048 has at least 1,024 tokens as context during prediction.
We report the loss breakdown by the type of document in \dataset.\looseness=-1

Figure~\ref{fig:dev_loss} shows that \model consistently outperforms other models. While this is expected and mainly serves as a sanity check, it also provides valuable insight into the generalization capabilities of the other models. For example, the gap to \model is most significant in the Filings category, likely because these documents, while public, are typically in PDF format and thus not included in any existing datasets.

\subsection{Financial Tasks}

\begin{table}[t]
\small
\centering
\begin{tabular}{@{}p{0.3\linewidth}@{} p{0.7\linewidth}@{}}
\toprule
Task & Template/Example \\ \midrule
\textbf{Discriminative} \\ \cmidrule{1-1}
Sentiment Analysis & \texttt{\{sentence\}\newline Question:~what is the sentiment?\newline Answer:~\{negative/neutral/positive\}} \\ 
Aspect Sentiment Analysis & \texttt{\{sentence\}\newline Question:~what is the sentiment on \{target\}?\newline Answer:~\{negative/neutral/positive\}} \\ 
Binary Classification & \texttt{\{sentence\}\newline Question:~\{question\}?\newline Answer:~\{Yes/No\}} \\
\textbf{Generative} \\ \cmidrule{1-1}
NER & Steve Jobs is the CEO of Apple\newline Extract named entity: Steve Jobs (person), Apple (organization)\\ 
NER+NED & AAPL stopped using Intel Chips\newline Extract ticker: AAPL, INTC\\ 
QA & \texttt{\{context\}\newline Question:~\{question\}?\newline Answer:~\{answer\}} \\
\bottomrule
\end{tabular}
\caption{Template for the different tasks we evaluate in the financial domain.}
\label{tab:eval-method}
\end{table}

The NLP tasks most often considered in finance are also common in the broader NLP literature; but, these tasks take on different characteristics and challenges when performed on financial data.
Take the example of sentiment analysis, where a headline such as ``COMPANY to cut 10,000 jobs'' portrays negative sentiment in the general sense but can at times be considered positive for financial sentiment towards COMPANY, as it might result in the stock price or investor confidence increasing.
We use a combination of public and internal benchmarks to assess the performance of \model, \evalcohort. All task types considered and their corresponding prompt templates are shown in Table~\ref{tab:eval-method}.\looseness=-1

\subsubsection{External Financial Tasks}\label{sec:fin-public}
Our public financial benchmarks include four tasks from the FLUE benchmark~\citep{shah-etal-2022-flue} and the ConvFinQA dataset~\citep{chen-etal-2022-convfinqa}.
As LLM performance on most of these financial tasks have not been broadly reported, there is no standard testing framework.
Thus, we adapt them to a few-shot setting (see Section~\cref{sec:method}).
Our guiding principle in designing the experiments was to select the number of shots such that the average performance across all the models was best.
While non-LLM numbers of custom models for these tasks are available, we omit reporting them here due to differences in the evaluation setup.
As a result, our claims are restricted to comparisons of LLMs.
We evaluate on the following tasks (more details provided in Appendix~\ref{app:external-tasks}):

\begin{itemize}
    \item \textbf{FPB}~\citep{DBLP:journals/jasis/MaloSKWT14}:
The Financial Phrasebank Dataset includes a sentiment classification task on sentences from financial news.
Any news that could benefit/hurt an investor is considered positive/negative and neutral otherwise.
We create our own splits and report F1 score weighted by support in a 5-shot setup.

\item \textbf{FiQA SA}~\citep{DBLP:conf/www/MaiaHFDMZB18}:
The second sentiment analysis task is to predict the aspect-specific sentiment in English financial news and microblog headlines, which were published as a part of the 2018 challenge on financial question answering and opinion mining.
While the original dataset is annotated on a continuous scale, we discretize the data into a classification setup with negative, neutral, and positive classes. Like with FPB, we create our own splits including microblogs and news, and use a 5-shot setup, reporting weighted F1.

\item \textbf{Headline}~\citep{DBLP:journals/corr/abs-2009-04202}:
This is a binary classification task of whether a news headline in the gold commodity domain includes certain information.
This human-annotated dataset consists of English news headlines about ``gold''.
Each news article carries a subset of the following tags: ``price or not'', ``price up'', ``price down'', ``price stable'', ``past price'', ``future price'', ``past general'', ``future general'', ``asset comparison''.
We verbalize each tag into a question using the official documentation, use 5 shots, and report the average weighted F1 score across all categories.

\item \textbf{NER}~\citep{salinas-alvarado-etal-2015-domain}:
This is a named entity recognition task on financial data gathered for credit risk assessment from financial agreements filed with the SEC.
The annotated entity types follow the standard CoNLL format~\citep{tjong-kim-sang-de-meulder-2003-introduction} and are annotated with PER, LOC, ORG, and MISC.
As it is nontrivial to learn to predict empty outputs in few-shot setups, we drop sentences that do not contain any entity. 
We further drop MISC tags due to their ambiguous definition.
All the models required more shots to perform well and we thus selected 20 shots and report the entity-level F1 score.

\item \textbf{ConvFinQA}~\citep{chen-etal-2022-convfinqa}:
Given input from S\&P 500 earnings reports that includes text and at least one table with financial data, the task is to answer conversational questions that require numerical reasoning over the input. 
This task requires numerical reasoning, an understanding of structured data and financial concepts, and a model needs to relate follow-up questions to the dialog turns.

For ConvFinQA, we use an entire gold conversation and its context is used as input to the models. As each ``turn'' of the conversation concludes, the ``turn'' along with the answer for that turn is appended as context for future turns.
We report the exact match accuracy on the public development set. 

\end{itemize}

\noindent
\model performs best of all models for four of the five tasks (ConvFinQA, FiQA SA, FPB, and Headline) and
comes in second in NER (\cref{tab:res-flue}).
Consequently, \model also has the highest win rate among all the models that we tested.
The gap to equally-sized models is especially pronounced for ConvFinQA which is challenging due to the requirement to use conversational input to reason over tables and generate an answer.\looseness=-1

\begin{table}[t]
\small
\centering
\scalebox{0.9}{
\begin{tabular}{l c c c c}
\toprule
 & \model & \gptneox & \opt & \bloombig{} \\
\midrule
ConvFinQA & \textbf{43.41} & 30.06 & 27.88 & 36.31 \\
FiQA SA & \textbf{75.07} & 50.59 & 51.60 & 53.12 \\
FPB & \textbf{51.07} & 44.64 & 48.67 & 50.25 \\
Headline & \textbf{82.20} & 73.22 & 79.41 & 76.51 \\
NER & 60.82 & \textbf{60.98} & 57.49 & 55.56 \\
\midrule
All Tasks \emph{(avg)} & \textbf{62.51} & 51.90 & 53.01 & 54.35 \\
All Tasks \emph{(WR)}  & \textbf{0.93} & 0.27 & 0.33 & 0.47  \\
\bottomrule
\end{tabular}
}
\caption{
    Results on financial domain tasks. 
}
\label{tab:res-flue}
\end{table}

\subsubsection{Internal Task: Sentiment Analysis}
For the Bloomberg-internal tasks, we consider aspect-specific sentiment analysis, which is prevalent in financial literature.
All of the datasets we use are in English.

Our annotation process consists of a discovery phase during which we establish the annotation and sampling procedures, understand how many annotators are typically required per example, and determine the level of training that is needed for the annotators~\citep{https://doi.org/10.13140/rg.2.2.34497.58727}. 
Depending on the complexity of the task, our annotators are a dedicated team of financial experts at Bloomberg, consultant workers, or a combination of both.
In each case, ties are resolved by adjudication from additional annotators and ambiguous examples are excluded.
All the datasets in this section were annotated by 2 annotators with a third annotator breaking any ties.

We measure the performance of LLMs for the internal datasets using a five-shot evaluation, similar to the external datasets. As the datasets are large, we randomly sample at most 1k test examples. We report F1 weighted by the support of each label.
Note that, similar to the external datasets, it is likely that the \emph{unlabeled} versions of the data used in our internal datasets occur in \dataset and are therefore seen by \model during training.
However, since some of \dataset is also available on the web, other LLMs we compare against may have also been trained on unlabeled versions of this data.
Dataset statistics are provided in ~\cref{tab:internal-sa-overview}.

\begin{table}[t]
\centering
\begin{tabular}{@{}llrrrrr@{}}
\toprule
Name & Time & Tokens & Test Size & \% Pos & \% Neu & \% Neg \\ \midrule
Equity News & 2018--2019 & 150-200 & 1,000 & 7 & 87 & 6 \\
Equity Social Media & 2015--2020 & 15-20 & 1,000 & 10 & 83 & 7 \\
Equity Transcript & 2008--2020 & 70-80 & 800 & 19 & 75 & 6 \\
ES News & 2016--2019 & 100-120 & 1,000 & 32 & 53 & 15 \\
Country News & 2009--2021 & 50-1,000 & 1,000 & 18 & 60 & 22 \\
\bottomrule
\end{tabular}
\caption{An overview of the Bloomberg-internal sentiment analysis tasks. Input token and label distribution numbers are computed on the test set.}
\label{tab:internal-sa-overview}
\vspace{-1em}
\end{table}

\begin{itemize}

\item \textbf{Equity News Sentiment}:
This task is to predict the aspect-specific sentiment expressed in the news story toward a company.
The dataset consists of English news stories from Bloomberg, premium, and web content.
Annotations of ``positive'', ``negative'', or ``neutral'' indicate that the news story is likely to increase, decrease, or not change the long-term investor confidence in the company.
\item \textbf{Equity Social Media Sentiment}:
The task is similar to  ``Equity News Sentiment'' but instead of news, we use financially-relevant English social media content.

\item \textbf{Equity Transcript Sentiment}:
This task is also similar to  ``Equity News Sentiment'' but instead of news, we use transcripts from company press conferences. %
The transcripts are made available through the use of speech recognition and at times, human edits.
Long transcripts are processed in chunks, and each chunk in our dataset typically contains between 70 and 80 tokens.

\item \textbf{ES News Sentiment}:
While this task is to predict the aspect-specific sentiment expressed in the news story towards a company (aspect), the goal is not to indicate effect on investor confidence.
The stories are annotated ``positive'', ``negative'', or ``neutral'' if the news story contains content that reflects good, bad, or neutral news about the company's environmental and social policies.
\item \textbf{Country News Sentiment}: 
This task is different from the other sentiment tasks in that the goal is to predict the sentiment expressed in the news story towards a country.
The dataset consists of English news stories from Bloomberg, premium, and web content. %
The stories are annotated ``positive'', ``negative'', or ``neutral'' if the news story alludes to the growth, shrinkage, or status quo of that country's economy.
\end{itemize}

~\cref{tab:res-internal-sa} shows that across the four internal aspect-specific sentiment tasks \model performs better than all the other tested models, by a wide margin. The only task in which the models perform similarly is the social media sentiment task, while \model outperforms the other models by at least 25 and up to over 60 points in the other three.

\begin{table}[t]
\small
\centering
\begin{tabular}{l c c c c}
\toprule
 & \model & \gptneox & \opt & \bloombig{} \\
 \midrule
Equity News & \textbf{79.63} & 14.17 & 20.98 & 19.96 \\
Equity Social Media & \textbf{72.40} & 66.48 & 71.36 & 68.04 \\
Equity Transcript & \textbf{65.06} & 25.08 & 37.58 & 34.82 \\
ES News & \textbf{46.12} & 26.99 & 31.44 & 28.07 \\
Country News & \textbf{49.14} & 13.45 & 17.41 & 16.06 \\
\midrule
All Tasks \emph{(avg)} & \textbf{62.47} & 29.23 & 35.76 & 33.39 \\
All Tasks \emph{(WR)} & \textbf{1.00} & 0.00 & 0.67 & 0.33 \\
\bottomrule
\end{tabular}
\caption{
Results on internal aspect-specific sentiment analysis datasets. \model far outperforms all other models on sentiment analysis tasks.
}
\vspace{-1em}
\label{tab:res-internal-sa}
\end{table}

\subsubsection{Exploratory Task: NER}

\begin{table}[t]
\small
\centering
\begin{tabular}{@{}llrrrr@{}}
\toprule
Name & Tokens & Test Size & LOC & ORG & PER \\
\midrule
BFW & $\roughly 21$ & 500 & 0.2 & 1.6 & 0.0 \\
BN & $\roughly 30$ & 500 & 0.7 & 1.0 & 0.6 \\
Filings & $\roughly 32$ & 500 & 0.1 & 1.3 & 0.4 \\
Headlines & $\roughly 50$ & 500 & 0.7 & 2.7 & 1.0 \\
Premium & $\roughly 29$ & 500 & 0.6 & 1.4 & 0.3 \\
Transcripts & $\roughly 23$ & 500 & 0.6 & 0.6 & 0.3 \\
Social Media & $\roughly 12$ & 500 & 0.4 & 1.4 & 0.2 \\
\bottomrule
\end{tabular}
\caption{An overview of statistics of our internal NER test set. We report average number of LOCation, ORGanization, PERson per example.}
\label{tab:internal-ner-overview}
\vspace{-1em}
\end{table}

Even though NER is a well-established NLP task with state-of-the-art results using BERT ~\cite{wu-dredze-2019-beto,luoma-pyysalo-2020-exploring} and T5~\cite{liu-etal-2022-autoregressive} style models, NER is largely an unexplored task for generative LLMs.
NER is not in HELM~\cite{HELM}, there is a single (Polish) task in BIG-bench~\cite{bigbench}, and none of the LLM papers we study report NER performance.
Hence, we consider NER as an exploratory task and report preliminary NER results given its importance in the Financial sector.

There are a few reasons for why NER may be a difficult task for generative LLMs.
NER is an information extraction task, and a better fit for encoder-decoder or encoder-only architectures.
The generative nature of LLMs does not confer an advantage for NER. We find that extensive prompt engineering and a greater number of shots are required to obtain reasonable results for NER than for other tasks.
Finance-specific NER has subtleties that make it especially difficult for zero or few-shot learning.

For example, consider the (fabricated) headline ``Bloomberg: Mr. Musk adds new features to Twitter and comments on China''.
Depending on our annotation guidelines and downstream task needs: (a) the reporting news organization ``Bloomberg'' can be tagged or not, depending on whether we want only salient entities, (b) ``Mr. Musk'' or just ``Musk'' is the PER to be tagged, (c) ``Twitter'' can be tagged as an ORG or a PRD (product) as features are added to the Twitter product and not the organization, and (d) ``China'' can be tagged ORG or LOC, though the right tag is likely ORG.
Without adding extensive annotation guidelines in the prompt, the LLM does not know the intended tagging behavior. 

Based on preliminary testing, we determined the following setting to obtain the best performance on the internal NER tasks from all models.
First, we restrict the entity types to be predicted to be ORG, PER, and LOC.
In all, we filtered out less than 1\% of entities.
We also remove all documents that contain no entities (i.e., all ``O'''s).
Both of these modifications are intended to increase the usefulness of the examples seen in few-shot prompting. We expect that further work on prompt engineering for NER could produce better results.

We consider seven Bloomberg internal NER datasets from different domains.

\begin{itemize}
\item \textbf{BN NER}:
This is a named entity recognition task on entities occurring in English long-form Bloomberg news content (the ``BN wire'') between 2017 to 2020.
\item \textbf{BFW NER}:
Similar to ``BN NER'' but instead of using the long-form BN wire, we use short-form stories from the ``Bloomberg First Word'' wire between 2018 to 2020.
\item \textbf{Filings NER}:
The goal of this task is to identify entities that occur in mandatory financial disclosures filed by companies.
The dataset contains filings sampled between 2016 and 2019.

\item \textbf{Headlines NER}:
The goal of this task is to identify entities that occur in headlines of English Bloomberg news content.
The dataset contains headlines sampled between 2016 and 2020.
\item \textbf{Premium NER}:
The goal of this task is to identify entities that occur in a subset of the third-party English news content ingested by Bloomberg.
The dataset contains stories sampled between 2019 and 2021.
\item \textbf{Transcripts NER}:
The goal of this task is to identify entities that occur in transcripts of company press conferences.
The dataset contains transcripts from 2019.
\item \textbf{Social Media NER}:
The goal of this task is to identify entities that occur in English financially-relevant social media content.
The dataset contains social media content sampled between 2009 and 2020.
\end{itemize}

\begin{table}[t]
\small
\centering
\begin{tabular}{l c c c c}
\toprule
 & \model & \gptneox & \opt & \bloombig{} \\
\midrule
\textbf{NER} \\ \cmidrule{1-1}
BFW & 72.04 & 71.66 & 72.53 & \textbf{76.87} \\
BN & 57.31 & 52.83 & 46.87 & \textbf{59.61} \\
Filings & 58.84 & 59.26 & 59.01 & \textbf{64.88} \\
Headlines & \textbf{53.61} & 47.70 & 46.21 & 52.17 \\
Premium & 60.49 & 59.39 & 57.56 & \textbf{61.61} \\
Transcripts & 75.50 & 70.62 & 72.53 & \textbf{77.80} \\
Social Media & 60.60 & 56.80 & 51.93 & \textbf{60.88} \\
\midrule
All Tasks \emph{(avg)} & 62.63 & 59.75 & 58.09 & \textbf{64.83} \\ 
All Tasks \emph{(WR)}  & 0.57 & 0.29 & 0.19 & \textbf{0.95}  \\
\midrule
\textbf{NER+NED} \\ \cmidrule{1-1}
BFW & \textbf{55.29} & 34.92 & 36.73 & 39.36 \\
BN & \textbf{60.09} & 44.71 & 54.60 & 49.85 \\
Filings & \textbf{66.67} & 31.70 & 65.63 & 42.93 \\
Headlines & \textbf{67.17} & 36.46 & 56.46 & 42.93 \\
Premium & \textbf{64.11} & 40.84 & 57.06 & 42.11 \\
Transcripts & \textbf{73.15} & 23.65 & 70.44 & 34.87 \\
Social Media & 67.34 & 62.57 & \textbf{70.57} & 65.94 \\
\midrule
All Tasks \emph{(avg)} & \textbf{64.83} & 39.26 & 58.79 & 45.43 \\
All Tasks \emph{(WR)} & \textbf{0.95} & 0.00 & 0.67 & 0.38 \\
\bottomrule
\end{tabular}
\caption{
Results on internal NER and NED datasets. On NER, while the much larger \bloombig{} model outperforms all other models, results from all models are relatively close, with \model outperforming the other two models. On NER+NED, \model outperforms all other models by a large margin.
}
\vspace{-1em}
\label{tab:res-internal-ner}
\end{table}

\noindent As our datasets are substantive, we randomly sample 4,000 training and 500 testing examples from each filtered internal dataset.
We utilize 20-shot prompts and evaluate using F1. 
The results from the internal NER tasks are mixed (\cref{tab:res-internal-ner}).
The much larger \bloombig{} wins most of the NER tasks.
Of the like-sized models, \model performs the best placing first once (Headlines), second four times (BN, Premium, Transcripts, Social media), third once (BFW), and last once (Filings).

\paragraph{Exploratory Task: NER+NED}
Named entity disambiguation (NED) links entity mentions to known entities in knowledge bases or other structured information sources. Within the financial world, we seek to link text mentions of companies to their ticker symbols, an abbreviation that uniquely identifies publicly traded shares of a particular stock on a particular stock market.

We directly test the ability of an LLM to complete this task by evaluating a joint NER+NED task: identify the stock tickers of companies mentioned in a document. This requires the model to first identify company mentions and then generate the corresponding stock ticker.
For example, 
given ``AAPL announced that they will stop using Intel chips in future products.'' the correct NER output would be ``AAPL, Intel'' while the correct NER+NED output would be ``AAPL, INTC''.

One of the advantages of this task is that it is robust to variations in extracting the exact text span. While NER evaluation requires exact matches, tickers may be successfully produced without first identifying spans. 
Furthermore, it evaluates a model's knowledge of companies, their various surface forms, and company to ticker mappings.

We create evaluation data with linked tickers for this task by running a state-of-the-art entity linking system for companies in financial data over the Bloomberg internal NER annotated documents from each domain. We remove documents with no linked tickers. Following our NER evaluations, we randomly sample 4,000 training and 500 testing examples from each filtered internal dataset. We utilize 20-shot prompts and evaluate using F1.

\cref{tab:res-internal-ner} shows that \model outperforms all other models by a large margin, except on social media data where it comes in second behind \bloombig. In our social media data, companies are often referenced by their tickers, removing the requirement of the model to link the mention and reverting the task to NER. These results further underscore the advantage of \model for financial tasks.

\subsection{BIG-bench Hard}
\label{sec:bbh}

\begin{table}[t]
\small
\centering
\scalebox{0.87}{
\begin{tabular}{l rrrr| r}
\toprule
BIG-bench Hard Task & \model & \gptneox & \opt & \bloombig{}& \palm \\
\midrule
Boolean Expressions$^{\lambda}$ & 62.40 & \textbf{71.20} & 48.40 & 69.20 & \textbf{83.2} \\
Causal Judgement & 49.73 & \textbf{52.41} & 51.87 & 51.87  & \textbf{61.0} \\
Date Understanding & \textbf{54.80} & 45.60 & 49.60 & 50.00   & 53.6 \\
Disambiguation QA & 34.00 & \textbf{40.80} & 40.40 & 40.40  &  \textbf{60.8} \\
Dyck Languages$^{\lambda}$ & 15.60 & 26.00 & 14.80 & \textbf{42.00} & 28.4 \\
Formal Fallacies & 50.80 & 52.80 & \textbf{54.00} & 52.80 &  53.6 \\
Geometric Shapes$^{\lambda}$ & 15.20 & 8.00 & 11.60 & \textbf{22.40} &  \textbf{37.6} \\
Hyperbaton & \textbf{92.00} & \textbf{92.00} & 91.60 & \textbf{92.00}  & 70.8 \\
Logical Deduction$^{\lambda}$ (\emph{avg}) & \textbf{34.53} & 30.93 & 31.87 & 34.00  & \bf{60.4} \\
Movie Recommendation & 90.40 & 86.40 & \textbf{91.20} & \textbf{91.20} & 87.2 \\
Multi-Step Arithmetic$^{\lambda}$ [Two] & \textbf{1.20} & 0.40	& 0.40 & 0.00 & \textbf{1.6} \\
Navigate$^{\lambda}$ & 42.00 & 45.20 & 42.00 & \textbf{50.00}  &  \textbf{62.4} \\
Object Counting$^{\lambda}$ & 33.20 & 21.20 & 26.00 & \textbf{36.80} &  \textbf{51.2} \\
Penguins in a Table & 37.67 & 33.56 & 28.08 & \textbf{40.41} &  \textbf{44.5} \\
Reasoning about Colored Objects & 34.80 & 26.00 & 31.20 & \textbf{36.80}   & \textbf{38.0} \\
Ruin Names & \textbf{56.00} & 54.00 & 52.80 & 54.80 &  \bf{76.0} \\
Salient Translation Error Detection & 20.00 & 20.40 & 16.40 & \textbf{23.60} & \textbf{48.8} \\
Snarks & 69.66 & 62.36 & 69.66 & \textbf{72.47}& \bf{78.1} \\
Sports Understanding & \textbf{62.80} & 53.20 & 54.40 & 53.20 & \textbf{80.4} \\
Temporal Sequences$^{\lambda}$ & 29.20 & 21.20 & 23.60 & \textbf{36.80} & \textbf{39.6} \\
Tracking Shuffled Objects$^{\lambda}$ (\emph{avg}) &\textbf{25.33} & 24.53 & 24.00 & 23.47 & 19.6 \\
Web of Lies$^{\lambda}$ & 49.20 & 52.40 & \textbf{54.00} & 51.20 &  51.2 \\
Word Sorting$^{\lambda}$ & 4.80 & 5.20 & 2.40 & \textbf{7.60} &  \textbf{32.0} \\
\midrule
NLP Task \emph{(avg)} & 54.39 & 51.63 & 52.60 & 54.96 & \textbf{62.7} \\
Algorithmic Task$^{\lambda}$ \emph{(avg)} & 28.42 & 27.84 & 25.37 & \textbf{33.95} & \textbf{40.9}  \\
\midrule
All Tasks \emph{(avg)} & 41.97 & 40.25 & 39.58 & \textbf{44.91} &  \textbf{52.3} \\
All Tasks \emph{(WR)} & 0.57 & 0.45 & 0.39 & \textbf{0.75}  & - \\
\bottomrule
\end{tabular}
}	
\caption{
    BIG-bench hard results using standard 3-shot prompting. Following the convention from \citet{BBH}, we denote algorithmic tasks with the superscript~$^{\lambda}$, and present averages for NLP and algorithmic categories. The baseline numbers from \palm{}~\citep{chowdhery2022palm} are taken from the original BBH paper. 
}
\vspace{-1em}
\label{tab:main_results_table}
\end{table}

We now turn to evaluate \model on standard, general-purpose NLP tasks. While the focus of our model is on financial tasks, our inclusion of general-purpose training data may help improve not only the financial tasks, but also allow our model to perform well on more standard NLP datasets.
We start with BIG-bench Hard~\citep{BBH}, a subset of the most challenging tasks in BIG-bench~\citep{bigbench}. It only includes tasks in which the best available model at construction was unable to achieve a performance higher than the average human rater via standard prompting techniques.

Results for each task are shown in \cref{tab:main_results_table}. Overall, while \model{} falls behind the much larger \palm{} (10x parameters) and \bloombig{} (3.5x parameters), it is the best-performing among similarly sized models. In fact, its performance is closer to \bloombig{} than it is to either \gptneox{} or \opt.
It further achieves the best performance of all models in date understanding, hyperbaton (ordering of adjectives), and tracking shuffled objects.
In sum, according to this benchmark, we find that developing finance-specific \model did not come at the expense of its general-purpose abilities.\looseness=-1

\subsection{Knowledge Assessments}
\label{sec:results-knowledge}

We next assess knowledge, which we define as the ability to recall information seen during model training, via scenarios that have the model answer questions without providing additional context or resources (closed-book question answering). This includes multiple-choice questions, and we report accuracy. We follow the template of \citet{GPT3}. The list of scenarios is as follows:

\begin{itemize}
  \setlength\itemsep{0.1em}
  \item \textbf{ARC}~\citep{Clark2018ThinkYH}: Multiple-choice questions collected from 3rd to 9th grade science exams, includes easy and challenging splits.
  \item \textbf{CommonsenseQA}~\citep{talmor-etal-2019-commonsenseqa}: Multiple-choice QA dataset that requires different types of commonsense knowledge.
  \item \textbf{MMLU}~\citep{MMLU}: Manually collected multiple-choice knowledge questions in 57 subjects.
  \item \textbf{PhysicalQA}~\citep[PiQA,][]{bisk2019piqa}: Questions about how the physical world works.
\end{itemize}

\begin{table}[t]
\small
\centering
\scalebox{0.9}{
\begin{tabular}{l rrrr|r}
\toprule
Task & \model & \gptneox & \opt & \bloombig & GPT-3 \\
\midrule
ARC (easy) &73.99&70.79&71.25&\textbf{75.93} & 71.2 \\
ARC (challenging) &48.63&45.39&44.54&\textbf{50.85}  & \textbf{53.2} \\
CommonsenseQA &65.52&60.36&\textbf{66.42} & 64.21 & - \\
PiQA & \textbf{77.86} &75.84&77.58 &77.04 & \textbf{80.5} \\ \midrule
All Tasks \emph{(avg)} & 66.50 & 63.10 & 64.95 & \textbf{67.01} & -  \\ 
All Tasks \emph{(WR)} &\textbf{0.75} & 0.08 & 0.33 & 0.67 & - \\ 

\bottomrule
\end{tabular}
}
\caption{
Knowledge tasks 1-shot results. 
The baseline numbers from GPT-3 are taken from~\citet{GPT3}. 
Among all models, \model achieves the highest win rate among the models we ran ourselves, and performs second best on average. 
}
\vspace{-1em}
\label{tab:res:knowledge}
\end{table}

\noindent \model achieves the highest performance among \evalcohort{} in one task, and comes second in the other three (\cref{tab:res:knowledge}). Similar to the previous section, it outperforms models of similar size while almost being on par with the much larger models.  
The Massive Multitask Language Understanding~\citep[MMLU,][]{MMLU} covers 57 different subjects and thus has a much wider coverage than the tasks described above. The aggregated results in \cref{tab:res-mmlu} paint a more consistent picture and follow the insights seen in BIG-bench hard. \model consistently outperforms \opt, which in turn outperforms \gptneox, while GPT-3 performs best. In contrast to the previous sections, \model also outperforms \bloombig{} in this category, although by a slim margin.
It falls behind the reported performance of GPT-3, especially in the social science category. The gap to GPT-3 is closest in the STEM and ``Other'' domains which include finance and accounting-related questions.

\begin{table}[t]
\small
\centering
\scalebox{0.9}{
\begin{tabular}{l cccc|c}
\toprule
Model & \model & \gptneox & \opt & \bloombig{} & GPT-3 \\
\midrule
Humanities & \textbf{36.26} & 32.75 & 33.28 & 34.05 & \textbf{40.8} \\
STEM & 35.12 & 33.43 & 30.72 & \textbf{36.75} & 36.7 \\
Social Sciences & 40.04 & 36.63 & 38.32 & \textbf{41.50} & \textbf{50.4} \\
Other & 46.36 & 42.29 & 42.63 & \textbf{46.48} & \textbf{48.8} \\
\midrule
Average & \textbf{39.18} & 35.95 & 35.99 & 39.13 & \textbf{43.9} \\
\bottomrule
\end{tabular}
}
\caption{Results (5-shot) on the MMLU~\citep{MMLU} benchmark. The baseline numbers from GPT-3 are taken from~\citet{MMLU}. While \model lacks behind \bloombig{} on three of the categories, its average is the highest among all models we evaluated ourselves. The gap to GPT-3 is largest on social sciences while the performance in other categories is close.}
\label{tab:res-mmlu}
\end{table}

\subsection{Reading Comprehension}
\label{sec:results-rc}

We define reading comprehension benchmarks as tasks in which the model can generate the correct response based on information contained in the presented input text. Our grouping includes open-book QA tasks, as opposed to \citet{GPT3}, who separate them into a different categories. 
We follow the template of \citet{GPT3}, and report accuracy. 
We include the following tasks:

\begin{itemize}
  \setlength\itemsep{0.1em}
  \item \textbf{BoolQ}~\citep{clark-etal-2019-boolq}: Yes/No questions about a passage from Wikipedia.
  \item \textbf{OpenBookQA}~\citep{mihaylov-etal-2018-suit}: Multiple-choice elementary-level science questions, given a book of science facts, applied to new situations.
  \item \textbf{RACE}~\citep{lai-etal-2017-race}: A multiple choice dataset of middle and high school English examinations.
  \item \textbf{Multi-Sentence Reading Comprehension} \citep[MultiRC,][]{khashabi-etal-2018-looking}: Short paragraphs and multi-sentence questions.
  \item \textbf{Reading Comprehension with Commonsense Reasoning} \citep[ReCoRD,][]{Zhang2018ReCoRDBT}: Automatically generated questions about CNN and Daily Mail news articles.
\end{itemize}

\begin{table}[t]
\small
\centering
\scalebox{0.9}{
\begin{tabular}{l c c c c | c}
\toprule
RC Scenario & \model & \gptneox & \opt & \bloombig & GPT-3 \\ \toprule
BoolQ &\textbf{74.59}&46.36&57.46&52.94&\textbf{76.7} \\ 
OpenBookQA &51.60&44.20&\textbf{58.00}& 47.20 &\textbf{58.8} \\ 
RACE (middle) &\textbf{54.32}&41.23&47.42&52.30 &\textbf{57.4}\\ 
RACE (high) &\textbf{41.74}&34.33&37.02&39.14 &\textbf{45.9}\\ 
MultiRC &\textbf{62.29} &22.86 &18.80 &26.65 &\textbf{72.9}\\ 
ReCoRD &\textbf{82.79}&	67.86&	82.53&78.01	&\textbf{90.2}\\ \midrule
All Tasks \emph{(avg)} & \textbf{61.22} & 42.81 & 50.21 & 49.37 & \textbf{67.0}  \\ 
All Tasks \emph{(WR)} & \textbf{0.94} & 0.06 & 0.50 & 0.50 & - \\ 
\bottomrule
\end{tabular}
}
\caption{Reading Comprehension Results (1-shot). The baseline numbers from GPT-3 are taken from~\citet{GPT3}. \model far outclasses the models we evaluated ourselves, and is slightly behind GPT-3.}
\vspace{-1em}
\label{tab:res-rc}
\end{table}

\noindent \cref{tab:res-rc} reflects a similar ranking as in the above evaluations: While GPT-3 has the highest performance, \model is a close second. Except for OpenBookQA, The performance of \model is the highest among \evalcohort. Surprisingly, \bloombig{} falls behind significantly in this category.

\subsection{Linguistic Tasks}
\label{sec:results-ling}

We define as linguistic tasks those scenarios that are not directly connected to user-facing applications. These include tasks that evaluate disambiguation, grammar, or entailment. These tasks are designed to directly assess a model's ability to understand language. We follow the template of \citet{GPT3}, and report accuracy. The list of tasks is as follows:

\begin{itemize}
  \setlength\itemsep{0.1em}
  \item \textbf{Recognizing Textual Entailment}~\citep[RTE,][]{Dagan2007ThePR,BarHaim2006TheSP,giampiccolo-etal-2007-third,DBLP:conf/tac/BentivogliMDDG09}: Given two text fragments, identify whether the meaning of one text is entailed.
  \item \textbf{Adversarial NLI}~\citep[ANLI,][]{nie-etal-2020-adversarial}: Adversarially constructed entailment detection.
  \item \textbf{CommitmentBank}~\citep[CB,][]{Marneffe2019TheCI}: Naturally occurring discourses whose final sentence contains a clause-embedding predicate.
  \item \textbf{Choice of Plausible Alternatives}~\citep[COPA,][]{Gordon2011SemEval2012T7}: Premise and two alternatives, where the task is to select the alternative that more plausibly has a causal relation with the premise.
  \item \textbf{Words in Context}~\citep[WIC][]{pilehvar-camacho-collados-2019-wic}: Determine if a word is being used with the same meaning in two sentences.
  \item \textbf{Winograd}~\citep{Levesque2011TheWS}: Determine which word a pronoun refers to when it is semantically unambiguous.
  \item \textbf{Winogrande}~\citep{Sakaguchi2019WINOGRANDEAA}: Adversarially mined challenging Winograd examples.
  \item \textbf{HellaSWAG}~\citep{zellers-etal-2019-hellaswag}: Pick the best ending to a story or set of instructions.
  \item \textbf{StoryCloze}~\citep{mostafazadeh-etal-2016-corpus}: Select the correct ending sentence for five-sentence long stories.
\end{itemize}

\begin{table}[t]
\small
\centering
\scalebox{0.9}{
\begin{tabular}{l rrrr|r}
\toprule
Linguistic Scenario & \model & \gptneox & \opt & \bloombig & GPT-3 \\ \toprule
RTE &\textbf{69.31} & 53.79 & 54.87 & 57.40 & \textbf{70.4}  \\
ANLI Round 1 &32.90 & 32.60 & 33.10 & \textbf{33.60} & 32.0 \\
ANLI Round 2 & \textbf{34.40} & 33.80 & 34.20 & 33.80 & 33.9 \\
ANLI Round 3 & \textbf{37.33} & 36.17 & 34.92 & 35.17 & 35.1 \\
\midrule
CB &\textbf{53.57} & 48.21 & 44.64 & 48.21 & \textbf{64.3} \\ 
COPA &86.00 &\textbf{88.00} &86.00 & 84.00 &\textbf{87.0} \\
WIC &\textbf{52.51} & 50.00 & \textbf{52.51} & 50.16 & 48.6 \\ 
\midrule 
WinoGrad &80.95 & 79.12 & \textbf{82.78} & 78.02 & \textbf{89.7} \\ 
WinoGrande &64.09 & 60.62 & 66.14 & \textbf{67.01} & \textbf{73.2} \\
\midrule
HellaSWAG &\textbf{73.92}&68.37&73.47&73.21 &\textbf{78.1} \\ 
StoryCloze &80.87&78.30&\textbf{81.83}&80.28 &\textbf{84.7}  \\ \midrule
All Tasks \emph{(avg)} & \textbf{60.63} & 57.18 & 58.59 & 58.26 & \textbf{63.4}  \\ 
All Tasks \emph{(WR)} & \textbf{0.85} & 0.27 & 0.58 & 0.42 & - \\ 
\bottomrule
\end{tabular}
}
\caption{
Results on the Linguistic Scenarios (1-shot).
The baseline numbers from GPT-3 are taken from~\citet{GPT3}. Win rates and averages are computed only based on accuracy numbers.
\model consistently scores highest among the models we evaluate, achieving an 85\% win rate. 
}
\vspace{-1em}
\label{tab:res-linguistic}
\end{table}

\noindent The results (\cref{tab:res-linguistic}) for linguistic tasks follow a similar trend to the knowledge category. \model falls slightly behind GPT-3 and outperforms the other models. Similar to the reading comprehension category, \bloombig{} falls behind \model.

\subsection{Summary} Across dozens of tasks in many benchmarks a clear picture emerges. Among the models with tens of billions of parameters that we compare to, \model performs the best. Furthermore, in some cases, it is competitive or exceeds the performance of much larger models (hundreds of billions of parameters). While our goal for \model was to be a best-in-class model for financial tasks, and we included general-purpose training data to support domain-specific training, the model has still attained abilities on general-purpose data that exceed similarly sized models, and in some cases match or outperform much larger models.

\section{Qualitative Samples}

We now share qualitative examples from our model that highlight the benefits of our domain specialization.

\paragraph{Generation of Bloomberg Query Language.}

One use case for \model is to make interactions with financial data more natural. An existing way to retrieve data is via the Bloomberg Query Language (BQL). BQL can be used to interact with different classes of securities, each with its own fields, functions, and parameters. BQL is an incredibly powerful but complex tool. As we show in Figure~\ref{fig:ex-bql}, \model can be utilized to make BQL more accessible by transforming natural language queries into valid BQL.

\begin{figure}[!t]
    \centering
    \begin{tcolorbox}[width=\textwidth,
                      interior hidden,
                      boxsep=0pt,
                      left=5pt,
                      right=0pt,
                      top=5pt,
                      ]%
\textbf{Input}: Get me the last price and market cap for Apple \\
\textbf{Output}: \verb|get(px_last,cur_mkt_cap) for(['AAPL US Equity'])| \vspace{8px} \\ 
\textbf{Input}: Tesla price \\
\textbf{Output}: \verb|get(px_last) for(['TSLA US Equity'])| \vspace{8px} \\ 
\textbf{Input}: Get the yield and spread for EC527035 Corp and AL580550 Corp \\
\textbf{Output}: \verb|get(yield,spread) for(['EC527035 Corp','AL580550 Corp'])| \vspace{8px} \\ 
\textbf{Input}: apple and ibm market cap and eps \\
\textbf{Output}: \verb|get(cur_mkt_cap,is_eps) for(['AAPL US Equity','IBM US Equity'])| \vspace{8px} \\ 
\textbf{Input}: industry subgroup of ibm apple microsoft google \\
\textbf{Output}: \verb|get(industry_subgroup()) for(['AAPL US Equity','IBM US Equity',|\\ \verb|'MSFT US Equity','GOOGL US Equity'])| 
\end{tcolorbox}
    \caption{Using \model to generate valid Bloomberg Query Language. Using only a few examples in a few-shot setting, the model can utilize its knowledge about stock tickers and financial terms to compose valid queries to retrieve the data, given a request in natural language. In each case, the model is given 3 examples (not shown) followed by the `Input'' and a prompt of ``Output:''.}
    \label{fig:ex-bql}
\end{figure}

\paragraph{Suggestion of News Headlines.}

Other use cases that are well supported are in the news space. Since it is trained on many news articles, it can be used for many news applications and assist journalists in their day-to-day work. 
For example, when constructing newsletters, journalists may have to write short headlines for each new section. 
While a dedicated model to help with this task may be too expensive to maintain, \model performs well out of the box (\cref{fig:ex-news}).

\begin{figure}[!t]
    \centering
    \begin{tcolorbox}[width=\textwidth,
                      interior hidden,
                      boxsep=0pt,
                      left=5pt,
                      right=5pt,
                      top=5pt,
                      ]%
\textbf{Input}:  The US housing market shrank in value by \$2.3 trillion, or 4.9\%, in the second half of 2022, according to Redfin. That's the largest drop in percentage terms since the 2008 housing crisis, when values slumped 5.8\% during the same period.
\\
\textbf{Output}: Home Prices See Biggest Drop in 15 Years \\
\\
\textbf{Input}:  The global economy is in a better place today than many predicted months ago, Janet Yellen said at the G20. At home, she pointed to a resilient US economy, where headline inflation has moderated and the labor market is strong. She also called for the IMF to move swiftly toward a fully-financed program for Ukraine.
\\
\textbf{Output}: Yellen Sees Global Economy More Resilient Than Expected \\
\\
\textbf{Input}:  Google was sued by the US and eight states seeking the breakup of its ad-tech business for allegedly monopolizing the digital advertising market. The lawsuit is the Biden administration's first big challenge to a tech titan and one of the rare times since 1982 that the DOJ has sought to cleave up a major company.
\\
\textbf{Output}: Google Sued for Monopoly in Online Ad Market 
\end{tcolorbox}
    \caption{Using \model to generate short headline suggestions in a three-shot setting. Bloomberg News sends many newsletters a day that requires these headlines. \model could help with the editing process by suggesting initial headlines from the text.}
    \label{fig:ex-news}
\end{figure}

\paragraph{Financial Question Answering.}

\begin{figure}[!t]
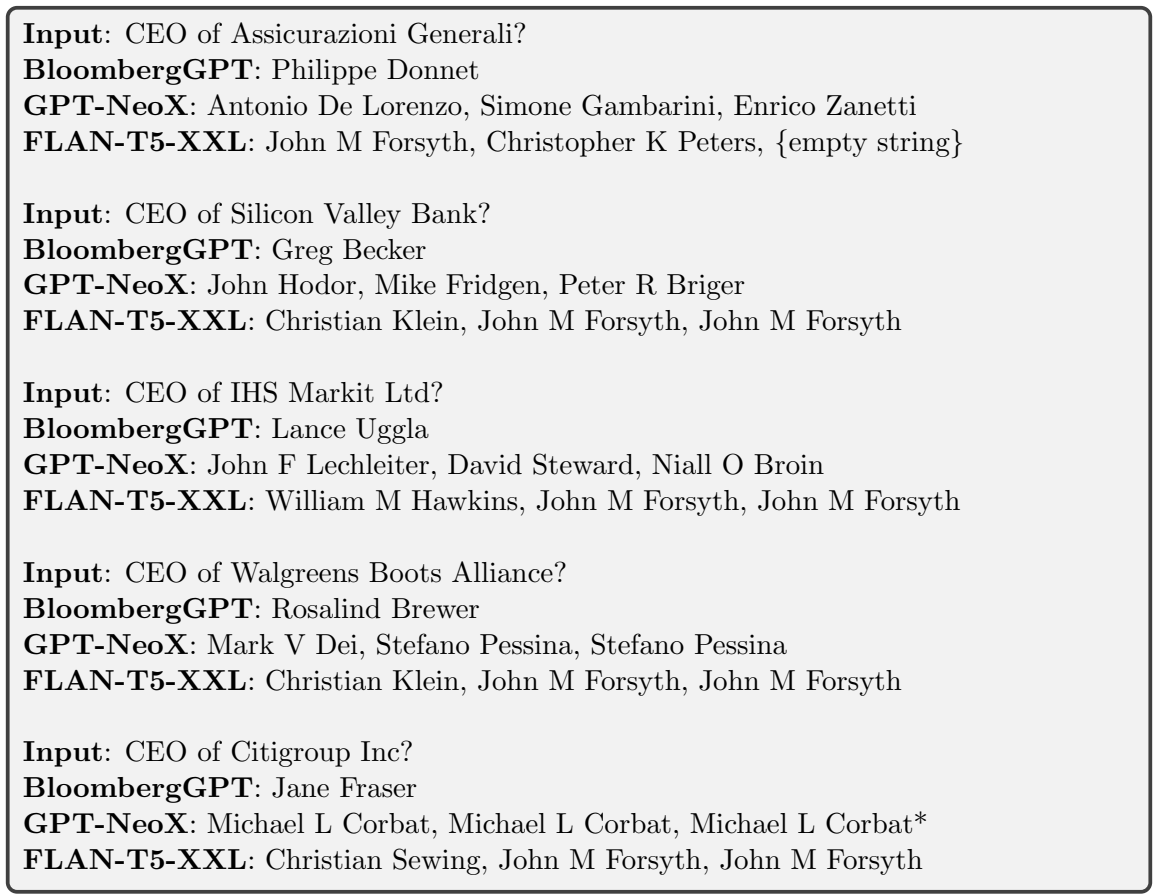

    \centering
    \begin{tcolorbox}[width=\textwidth,
                      interior hidden,
                      boxsep=0pt,
                      left=5pt,
                      right=5pt,
                      top=5pt,
                      ]%
\textbf{Input}:  CEO of Assicurazioni Generali?
\\
\textbf{\model}:  Philippe Donnet\\
\textbf{\gptneox}: Antonio De Lorenzo, Simone Gambarini, Enrico Zanetti \\
\textbf{FLAN-T5-XXL}: John M Forsyth, Christopher K Peters, \{empty string\} \\
\\
\textbf{Input}: CEO of Silicon Valley Bank? 
\\
\textbf{\model}: Greg Becker \\
\textbf{\gptneox}: John Hodor, Mike Fridgen, Peter R Briger \\
\textbf{FLAN-T5-XXL}: Christian Klein, John M Forsyth, John M Forsyth \\
\\
\textbf{Input}: CEO of IHS Markit Ltd?
\\
\textbf{\model}: Lance Uggla \\
\textbf{\gptneox}: John F Lechleiter, David Steward, Niall O Broin \\
\textbf{FLAN-T5-XXL}: William M Hawkins, John M Forsyth, John M Forsyth \\
\\
\textbf{Input}: CEO of Walgreens Boots Alliance? 
\\
\textbf{\model}: Rosalind Brewer \\
\textbf{\gptneox}: Mark V Dei, Stefano Pessina, Stefano Pessina \\
\textbf{FLAN-T5-XXL}: Christian Klein, John M Forsyth, John M Forsyth \\
\\
\textbf{Input}:  CEO of Citigroup Inc?
\\
\textbf{\model}: Jane Fraser  \\
\textbf{\gptneox}: Michael L Corbat, Michael L Corbat, Michael L Corbat* \\
\textbf{FLAN-T5-XXL}: Christian Sewing, John M Forsyth, John M Forsyth 
\end{tcolorbox}
    \caption{Testing the ability of \model, \gptneox, and FLAN-T5-XXL to recall the names of CEOs of companies. Each model is run in a 10-shot setting. We sample up to three answers and present all of them if they are incorrect. *Michael Corbat was CEO of Citigroup until 2021, highlighting the importance of an up-to-date model.}
    \label{fig:ex-ceo}
\end{figure}

Due to the financial domain training data, we are able to query \model for knowledge relevant to the financial world. For example, it performs well at identifying the CEO of a company. \cref{fig:ex-ceo} shows several examples including output from other models. 
While \model correctly identifies the CEOs, \gptneox{} does not, and FLAN-T5-XXL completely fails, consistently ignoring the company and instead predicting the CEO at Cirrus Logic who was included in the prompt. 
While \model{} does not perfectly solve this task and makes mistakes, we were not able to find any example where the other models solved the task while \model did not.

\section{Related Work}
\label{sec:related_works}
\paragraph{Language Models.}
Language modeling has a long history in the NLP community. The idea of training a probabilistic language model for scoring word sequences was likely first introduced by \citet{jelinek1976continuous}.
N-gram models were popular for decades \cite{brown1992class}, and were trained on corpora up to 2 trillion tokens \citep{brants-etal-2007-large}.
Research on training language models accelerated over the
last decade due to innovations in machine learning, data availability, and
compute.
Early work in autoregressive language modeling \citep[e.g.,][]{mikolov2010recurrent,
sutskever2011generating} used recurrent neural networks, but these were small models trained on small datasets.
The introduction of the transformer architecture
\citep{vaswani2017attention} facilitated the scaling of these models in terms of data, compute, and the number of parameters.

The process of developing models that could better approximate the distribution of language over large corpora led to the discovery that the representations these models produce are useful starting points for many downstream tasks. 
This was demonstrated by \citet{gpt-radford} and \citet{howard-ruder-2018-universal} who showed that generative pretraining with an autoregressive language modeling objective achieves strong performance in transfer learning.
\citet{gpt-2-radford} further showed scaling the model size and
training data led to autoregressive language models that perform well in different downstream tasks without any additional supervised fine-tuning.

\citet{GPT3} showed that further scaling the models led to the emergence of new model capabilities and increased model robustness.
Since the release of GPT-3 by \citet{GPT3}, many other researchers built large language models to study data quantity, data quality, network architecture, parameter scaling, data scaling, tokenization, and open-sourcing strategies~\citep[][among many others]{JMLR:v21:20-074, opt-zhang, black-etal-2022-gpt,
gopher-rae, hoffmann2022chinchilla, chowdhery2022palm,
lieber2021jurassic, glm-zeng, tafjord2021general,
https://doi.org/10.48550/arxiv.2201.11990, bloom, galactica-taylor,
lin2022language, soltan2022alexatm, thoppilan2022lamda, bao-etal-2022-plato,
sanh2022multitask, roller-etal-2021-recipes, glaese2022improving, wang-etal-2021-codet5,
peng2022godel}.

\paragraph{Domain-Specific Large Language Models.}
The value of domain-specific training for masked (encoder only) language models is well established. Commonly accepted approaches are to train BERT models~\citep{devlin-etal-2019-bert} from scratch on domain-specific data or to continue pretraining an existing model on new domain-specific data~\citep{gururangan-etal-2020-dont}.
Following these strategies, BioBERT~\citep{biobert-lee} adapts BERT to the biomedical domain and SciBERT is trained on scientific publications~\citep{beltagy-etal-2019-scibert}.
The results of these papers showed that in-domain training allows models to outperform previous state-of-the-art models in a variety of biomedical text mining tasks.
Further examples of this paradigm are ClinicalBERT for the clinical domain~\citep{clinicalbert-huang}, BioMed-RoBERTa for scientific biomedical papers~\citep{gururangan-etal-2020-dont}, and BERTweet and Bernice for Twitter data~\citep{nguyen-etal-2020-bertweet,delucia-etal-2022-bernice}. 

Since the training of auto-regressive---decoder-only---language models of more than 10B parameters is significantly more costly than training masked LMs under 1B parameters, there have been much fewer examples of domain-specific autoregressive models.
However, existing approaches follow the same two strategies. 
Adapting an existing model, medPaLM \citep{medpalm} adapted PaLM to the biomedical domain and Minerva \citep{minerva} to mathematical reasoning tasks. 

Recently, several examples of from-scratch trained decoder-only models for domain-specific data have emerged. One popular domain is protein sequences since they can be represented using language-like sequences but are not covered by natural language models~\citep[e.g.,][]{lin2022language,DBLP:journals/corr/abs-2108-07435,DBLP:journals/corr/abs-2206-13517}.
However, there can be benefits even for models in natural language domains. Galactica is trained exclusively on a large collection of scientific datasets, and includes special processing to handle scientific notations \citep{galactica-taylor}.
While performing very well on scientific tasks, Galactica also surprisingly also performs well on more standard NLP tasks.
BioGPT \citep{Luo_2022} and BioMedLM \cite{biomedlm} are both smaller GPT-style models trained on biomedical data. \citet{lehman-clinicalt5} compares encoder/decoder models trained exclusively on domain-specific data, versus those adapted from general-purpose training.
Researchers working on large generative language
dialog models have reached similar conclusions about the benefits of using domain-specific training data \citep{zhang-etal-2020-dialogpt, roller-etal-2021-recipes, thoppilan2022lamda}.

These findings highlight the advantages of in-domain pretraining, especially if sufficient data is available, as it is in our case. Inspired by the general capabilities of Galactica, we augment our private data with public data with the goal of investigating whether a model can gain in-domain capabilities without sacrificing general-domain performance. 

\paragraph{Training Data.}
Large corpora of raw text data are critical for training LLMs.
As a result, there are now several corpora available that cover a wide range of sources.

The Colossal Clean Crawled Corpus~\citep[C4,][]{JMLR:v21:20-074} 
draws from Common Crawl to create a processed training corpus.
The Pile is a carefully curated corpus that contains a wide range of data sources \cite{the-pile}.
These datasets are built on or include web crawls (OpenWebText2) augmented with an array of data from high-quality sources (Pubmed, Arxiv).
Various efforts aim to clean datasets, especially web data, by removing unwanted or harmful text~\citep{llama,gopher}. BLOOM \cite{bloom} carefully selected data sources and included various filtering mechanisms \cite{bloom-data-governance}.

While web data is an effective strategy for obtaining large amounts of diverse data, robust cleaning efforts still result in data artifacts, duplicates \cite{carlini2021extracting}, various types of toxic language \cite{welbl-etal-2021-challenges-detoxifying}, and it can lead to unintended marginalization of minority voices~\citep{xu-etal-2021-detoxifying}.
\citet{dodge-etal-2021-documenting} studied C4 to better understand the metadata, and the included and excluded data.
Their findings suggest that C4 contains machine-generated text,
is biased due to exclusion filters and might contain examples drawn from
evaluation datasets for NLP tasks. A similar effort was undertaken by \citet{glm-zeng} to document the pre-processing they undertook to
train their Chinese large language model.

\citet{lee-etal-2022-deduplicating} investigated the impact of deduplication on model performance for several datasets and found that deduplication reduces the emission of memorized training data, allows better estimation of the generalization error, and improves training time and cost without impacting performance.
These insights highlight the importance and challenges of constructing high-quality training corpora.
As discussed in \S\ref{sec:dataset}, Bloomberg's core business curates and provides access to datasets, which we use to construct a high-quality dataset \dataset to train \model, resulting in best-in-class financial performance.

\paragraph{Evaluation.} The tasks addressed by language models have vastly increased and require a very different evaluation process from traditional task-specific systems. There have been two paradigms for LLM evaluation: The first is to evaluate a model in many different scenarios via automatic evaluation~\citep{HELM,bigbench} and the second is to perform extrinsic and task-specific evaluations by integrating them into user workflows~\citep[e.g.,][]{DBLP:journals/corr/abs-2212-09746,https://doi.org/10.48550/arxiv.2209.12356}. 

While the second strategy is necessary for assessing deployments of models in products, it is infeasible to run these human evaluations at a scale of the first strategy and it is thus standard to follow the first strategy when introducing new models. In our case, we combine multiple general-purpose evaluations from multiple existing benchmarks that have different goals. \citet{bigbench} aim for maximum coverage by soliciting tasks from the entire research community, while HELM~\citep{HELM} suggests evaluation in various ``scenarios'' that are represented through specific datasets. Earlier language model papers developed their own evaluation schemata~\citep{GPT3}. While these benchmarks allow for a side-by-side comparison between models, it is challenging to ensure that all experimental parameters (prompts, decoding strategies, few-shot examples, etc.) are the same. For that reason, we differentiate between reported and verified numbers in our evaluation (\S\ref{sec:eval}).

Beyond the general-purpose evaluation, we also require a targeted domain evaluation. Prior domain-specific models like Galactica~\citep{galactica-taylor} chose a set of tasks that the model is likely to perform well on. In their case, these were various scientific tasks. However, there exists no standard benchmark for the financial NLP domain. While the recent work on FLUE~\citep{shah-etal-2022-flue} aims to provide such a benchmark, it has limited coverage of relevant tasks, no suggested evaluation strategy for few-shot learning, and the quality of some annotations is low. To provide externally comparable results, we developed a few-shot strategy for FLUE, but also decided to augment the publicly available evaluation tasks with company-internal benchmarks. 

\paragraph{Model Size.}
Large language model training remains expensive in terms of the computational cost and human effort to assemble data and train the model. Determining the optimal amount of training data and model shape and size for the best utilization of resources becomes important.

\citet{scaling-kaplan} first studied the dependence of language model performance
on architecture, parameter size, compute power, and dataset size.
They reported that the number of model parameters, the dataset size, and the
amount of compute improves performance on the autoregressive language modeling
objective smoothly according to the power law.
A similar investigation by \citet{hernandez2021scaling} into data
transfer for differing distributions found that this also follows a power law.
Moving beyond studying the effect on loss, \citet{gopher-rae} analyzed the effect of scale on undesirable properties such as bias and toxicity by training a wide range of model sizes.

Comparing model architectures, \citet{levine2020limits} studied the scaling of models that use self-attention and derived guidelines for depth-to-width allocation.
\citet{tay2021scale} reported that model shape (depth-width
ratio) impacted performance on downstream tasks even if it had minimal
impact on the pretraining objective.
\citet{tay2022scaling} further studied the effect of scaling for
different model architectures and showed that architecture choice is pertinent
when scaling and that the vanilla transformer architecture scales  best.

Of particular importance to this work is the study of 
\citet{hoffmann2022chinchilla}, who investigated the effect of model size and
the number of training tokens on the performance of a model given a fixed compute budget.
They posited that existing large language models were undertrained and that model size and the number of training tokens should be scaled equally.
They demonstrated this hypothesis through Chinchilla, a model significantly smaller, yet higher performing, than most of the largest LLMs.
These findings opened the door for ``Chinchilla optimal'' training of smaller models that achieve strong performance, and for which inference can be run much more efficiently than for their larger counterparts.
These findings led us to consider a nearly Chinchilla-optimal model using a standard architecture.

\paragraph{Tokenization.}
Tokenization and vocabulary choice play a critical role in model performance as they can help the model learn meaningful representations and generalize to unseen
words.
Byte-Pair encoding (BPE) \citep{sennrich-etal-2016-neural} learns a greedy bottom-up
vocabulary by repeatedly merging the most frequent sequence pairs in the training set till a predetermined vocabulary size is reached.
\citet{gpt-radford} adapted BPE by limiting the base vocabulary to be all
possible bytes as opposed to all Unicode characters.
Wordpiece tokenization \citep{schuster2012japanese} also learns a greedy
bottom-up vocabulary by repeatedly merging the sequence-pair that maximizes the
likelihood of the training data, which is a slight deviation from the method in
\citet{sennrich-etal-2016-neural}.

In contrast to BPE and Wordpiece, the Unigram tokenizer
\citep{kudo-2018-subword} learns a top-down vocabulary by first initializing
a large vocabulary and repeatedly discarding those vocabulary items that
increase loss (e.g., log-likelihood of the training data) the least.
By construction, the Unigram model can tokenize an input text in several
different ways.
That is, the Unigram model saves probabilities allowing for smarter tokenization at inference time.\looseness=-1

Finally, SentencePiece \citep{kudo-richardson-2018-sentencepiece} adapts the schemes
mentioned above to handle languages that are not space separated.
\citet{beltagy-etal-2019-scibert} constructed a vocabulary specific to scientific
text and observed that their domain-specific trained vocabulary only had a 42\%
overlap with the non-domain-specific BERT vocabulary trained on general domain
text.
Similarly, \citet{lewis-etal-2020-pretrained} showed that a dedicated biomedical
vocabulary improved performance on sequence labeling tasks consistently.
\citet{lieber2021jurassic} constructed a larger vocabulary to ensure token
efficiency, which the authors claim resulted in reduced training time and
better semantic representation.
These findings demonstrate the importance of selecting a tokenizer and accompanying vocabulary that best reflects that training domain. 
For those reasons, we decided to train our own unigram tokenizer instead of relying on existing public ones.

\paragraph{Positional Embeddings.}
Transformer-based models rely on positional embeddings to encode position and location information of words in a text.
Encoding the sequence position and the effect of this choice on model
performance have been studied extensively. These include sinusoidal embeddings
\citep{vaswani2017attention}, rotary position embeddings
\citep{su2021roformer}, adding relative position bias \citep{JMLR:v21:20-074},
and adding linear biases to attention heads \citep{press2022train}.
A side-effect of the strategy in \citet{press2022train} is that one can train
on shorter sequences without loss in performance on longer sequences.
This has two benefits: first, models learn to generalize (extrapolate)
to longer sequences and second, models can be trained on shorter sequences
reducing training time.

\section{Ethics, Limitations, and Implications}

The rapid development and adoption of large language models have been accompanied by a rigorous conversation about the ethics, uses, and limitations of these models. For a more complete treatment of these topics, we direct the reader to \citet{bommasani2021opportunities,bender2021dangers,birhane2022values,https://doi.org/10.48550/arxiv.2112.04359,weidinger2022taxonomy}.
We discuss issues that are directly relevant to the development of \model.

\subsection{Ethical Use}

Finance is a sensitive area for technology, and ensuring accurate, factual information is crucial for our products, our clients, and the firm's reputation in the marketplace.
On the other hand, our clients are also eager to adopt state-of-the-art technology to support their workflows. To provide natural language applications to the financial community, we have developed a rigorous risk and testing assessment process. This process includes careful annotation guidelines \cite{https://doi.org/10.13140/rg.2.2.34497.58727}, pre-launch review at multiple levels by the central risk and compliance organizations, and by the product leaders (e.g., the newsroom) as applicable, and post-launch monitoring. Moreover, we conduct our  research, development, and deployment of NLP and AI systems in accordance with all applicable regulations.

Similarly, toxicity and bias are areas where, as a company, we take extraordinary care with any content we produce, whether from humans or machines. Since the measurement of toxicity and bias in our model depends on its application areas, quantifying the potential for the generation of harmful language remains an open question. We are particularly interested in studying whether \dataset, which is cleaner and contains fewer examples of overtly biased or toxic language (e.g., Press Releases), reduces the proclivity of the model to generate inappropriate content. %
As we move to develop products built on this technology, we will apply existing testing procedures, as well as risk and compliance controls, to ensure safe use.\looseness=-1

\subsection{Openness}
An ongoing debate in the community concerns how LLMs should be released, if at all. While models that are not publicly available cannot be fully evaluated by the community, distributing models can lead to nefarious purposes. Especially for a model like \model, which is trained on a significant amount of press releases, news articles, and filings, a release carries a high risk for abuse through imitation. 

We have witnessed many different strategies to mitigate risks associated with the release of LLMs.
One strategy is to freely and openly share trained models \cite{bloom}, and rely on a license that dictates how the model should and should not be used. 
Another requires individuals to apply for access to the trained model parameters \cite{opt-zhang,llama}. 
A more restrictive approach is to provide API access to models, but no access to the underlying model parameters or detailed information on the data the model was trained on~\citep{GPT3}. 
Finally, some have provided no access to the model \cite{chowdhery2022palm,hoffmann2022chinchilla}. Each decision reflects a combination of factors, including model use, potential harms, and business decisions.

One of Bloomberg's core business propositions is around providing access to data that has been collected over the course of decades. As is well known, LLMs are susceptible to data leakage attacks and it is possible to extract significant segments of text given model weights \cite{carlini2021extracting,https://doi.org/10.48550/arxiv.2202.07646}. Moreover, even giving selective access to researchers isn't a guarantee that the model cannot be leaked. Without strong privacy guarantees, we must be concerned that providing access to model weights entails giving access to \dataset.
For this reason, we err on the side of caution and follow the practice of other LLM developers in not releasing our model.

Nevertheless, our insights and experiences in training and evaluating \model contribute to the developing understanding of these models. In particular, our experience may be useful to those building domain-specific models. During the process of developing \model, we found the OPT chronicles, experiences of the BLOOM team, as well as work of non-open models like GPT-3, PaLM, Chinchilla, and Gopher, to be crucial enablers of our work. In support of this tradition, we include our Training Chronicles (\cref{sec:chron}).

\section{Conclusion}
We have presented \model, a best-in-class LLM for financial NLP.

Our model contributes to the ongoing dialog on effective ways to train domain-specific models. Our training strategy of mixing domain-specific and general-purpose data results in a model that balances performance in both domains. Additionally, our work offers another data point on selecting Chinchilla optimal-sized models. Finally, we hope that our model training logs will provide a guide for those training their own LLMs.

We have several interesting directions to pursue. First, task fine-tuning has yielded significant improvements in LLMs, and we plan to consider what unique opportunities exist for model alignment in the financial domain \citep{flan-paper,ouyang2022InstructGPT}. Second, 
by training on data in \dataset, we are selecting data that may exhibit less toxic and biased language. The effects of this on the final model are unknown as yet, which we plan to test. Third, we seek to understand how our tokenization strategy changes the resulting model. These are a few of the new research directions we hope to pursue with \model.

We achieve strong results on general LLM benchmarks and outperform comparable models on financial tasks.
We attribute this, in decreasing order of impact, to 1. a well-curated internal dataset, 2. our unique choice in tokenizer, and 3. an up-to-date architecture. We will continue to develop financial applications with \model to further explore the benefits of these modeling choices.

\acks{We would like to acknowledge the people who helped us, including Emmanuel Scoullos (NVIDIA) and Can Karakus (Amazon Web Services).}

\bibliography{refs, anthology}
\newpage

\appendix

\newcommand{\softmax}{\mathop{\mathrm{softmax}}\nolimits}
\newcommand{\drop}{\mathop{\mathrm{drop}}\nolimits}
\newcommand{\gelu}{\mathop{\mathrm{gelu}}\nolimits}
\newcommand{\modop}{\mathop{\mathrm{mod}}\nolimits}

\newcommand{\vect}[1]{\boldsymbol{#1}}
\newcommand{\mat}[1]{\boldsymbol{#1}}

\definecolor{paramc}{RGB}{225, 102, 102}
\newcommand{\param}[1]{\textcolor{paramc}{#1}}

\section{Architecture}
\label{sec:architecture}

\setcounter{subsection}{-1}

\subsection{Notation}

\paragraph{Styling.} Unstyled variables denote scalars, bold lower-case variables represent [column] vectors, and bold capitalized variables represent matrices. For instance, $h_{i,j}$ could be an element in the vector $\vect{h_j}$, which could in turn be the $j$-th column of matrix $\mat{H}$.

Named functions are typed in non-italicized regular typeface, such as $\softmax(\cdot)$ and $\FFN(\cdot)$.

\param{Red color} is used to denote trainable parameters, or functions that are parametrized by trainable parameters, such as $\param{\mat{W}}$ or $\param{\FFN(} \cdot \param{)}$.

\paragraph{Sequences.} A sequence $(x_1, \mathellipsis, x_n)$ of $n$ elements is denoted by $\{x_i\}_{i=1}^n$. We treat a sequence of (column) vectors as a matrix, i.e. $\mat{X} = \{ \vect{x_i} \}_{i=1}^n \in \mathbb{R}^{m\times n}$, where each $\vect{x_i}\in \mathbb{R}^{m}$.

\paragraph{Operators.}
\begin{itemize}
\item $f: \mathbb{R}^n \rightarrow \mathbb{R}^n$: A function on vectors, that is, $\vect{y} = f(\vect{x})$ where $\vect{x}, \vect{y} \in \mathbb{R}^n$ are $n$-dimensional real valued vectors. Whenever such a function is applied to a matrix, it is applied column-wise: $f(\mat{X}) = \{ f(\vect{x_j}) \}_{j=1}^m$, $\mat{X} \in \mathbb{R}^{n \times m}$.

\item $\mat{A} \odot \mat{B}$: Element-wise (or Hadamard) product of matrices or vectors $\mat{A}$ and $\mat{B}$ (of the same shape).

\item $\mathbbm{1}(P)$: Indicator function that returns 1 if the predicate $P$ is true and 0 otherwise.

\item $[n]$: For integer $n$, the set of all positive integers up to (including) $n$, i.e. $\{1,\mathellipsis,n\}$.

\item $\mat{A} + \vect{b}$: Adding a vector to a matrix is defined as repeated addition to each column. That is, $\mat{A} + \vect{b} = \{\vect{a_i} + \vect{b}\}_{i=1}^n$.
\item Softmax: $\softmax(\vect{x}) = \dfrac{\exp(\vect{x})}{\sum_i^n \exp(x_i)}$ where $\exp(\cdot)$ is applied element-wise to a vector.
\item Dropout: $\drop^p(\vect{x}) = \dfrac{1}{1-p} \cdot \vect{m} \odot \vect{x}$ where, $\vect{m} = {[m_i]_{i=1}^n}^\top$, and $m_i \sim \text{Bernoulli}(1-p)$. Random variables $m_i$ are drawn independently for each presentation of an example. %
\end{itemize}

\subsection{Full Architecture}

\paragraph{Embedding.} Let $(x_1, \mathellipsis, x_t) = \{x_t\}_{t=1}^T \in \mathcal{V}^T$ denote an input sequence of length $T$, where each element $x_t$ denotes an integer identifier of a token from the vocabulary $\mathcal{V} = [|\mathcal{V}|]$.
 
Initial input representations $\mat{H^0} = \{\vect{h_t^0}\}_{t=1}^T$ are obtained by 
\begin{align}
    \vect{\bar{h}_t^0} &= \param{\mat{W^{em}}} \vect{e_{x_t}} & \forall t\\
    \vect{h_t^0} &= \param{\LN^{em}(}\vect{\bar{h}_t^0}\param{)} & \forall t
\end{align}
where $\param{\mat{W^{em}}} \in \mathbb{R}^{D \times |\mathcal{V}|}$ is the token embedding matrix, $\vect{e_{x_t}} \in \mathbb{R}^{|\mathcal{V}|}$ is the $x_t$-th standard basis vector, and $\param{\LN^{em}}$ is the \emph{em}bedding LayerNorm function, to be defined in the following subsections.

Observe that no positional embedding is applied here due to how ALiBi works.

\paragraph{Layers.} Layer representations $\mat{H^\ell} \in \mathbb{R}^{D \times T}$ for each layer $\ell = 1, \mathellipsis, L$ can be sequentially defined as follows (this computation is sometimes referred to as a ``block''):
\begin{align}
\mat{\bar{H}^\ell} &= \mat{H^{\ell - 1}} + \param{\SA_\ell(\LN^{in}_\ell(}\mat{H^{\ell-1}}\param{))} & \forall \ell \\ %
\mat{H^\ell} &= \mat{\bar{H}^\ell} + \param{\FFN_\ell( \LN^{at}_\ell (} \mat{\bar{H}^\ell} \param{) )} & \forall \ell
\end{align}
where $\param{\SA_\ell}$, $\param{\FFN_\ell}$, and $\param{\LN_\ell^{\bigcdot}}$ denote 
SelfAttention, FeedForwardNetwork, and LayerNorm functions at layer $\ell$, respectively, as 
defined in the following subsections. The red color indicates that the functions depend on trainable parameters. $\param{\LN_\ell^{\bigcdot}}$ is further parametrized by an indication of what the function is applied to, such as $\param{\LN_\ell^{in}}$ when applied to the block 
\emph{in}put and $\param{\LN_\ell^{at}}$ when applied to the \emph{at}tention output. We designate these separately since they use different (i.e. untied) trainable parameters.

\paragraph{Logits.} Given the final layer representation $\mat{H^L}$, logits $\mat{Y} \in \mathbb{R}^{|\mathcal{V}|\times T}$ are obtained as:
\begin{align}
    \mat{Y} &= {\param{\mat{W^{em}}}}^\top \param{\LN^f(}\mat{H^L} \param{)}
\end{align}
where  $\param{\mat{W^{em}}} \in \mathbb{R}^{D \times |\mathcal{V}|}$ is the same embedding matrix we used in the {\bf embedding} part and $\param{\LN^f}$ is the \emph{f}inal LayerNorm application.
We follow the PaLM approach in omitting a bias term. 

The token distribution for position $j+1$, conditioned on the prefix $(x_1,\dots,x_j)$, is given by
\begin{align}
    \mathbb{P}(x_{j+1} = w | \{x_t\}_{t=1}^j) = \softmax (\vect{y_j})_w \label{eq:lm-head}
\end{align}
where $\vect{y_j}$ is the $j$'th column of $\mat{Y}$.
\subsection{SelfAttention with ALiBi (\texorpdfstring{$\SA$}{SA})}

SelfAttention with ALiBi at layer $\ell$, $\param{\SA_\ell}: \mathbb{R}^{D \times T} \rightarrow \mathbb{R}^{D \times T}$ is defined as follows.

Let $n \in \{1, \mathellipsis, N\}$ %
denote an attention head where $N$ is the total number of heads. Let $D^n$ denote the dimensionality of each head. %
Let $\mat{A^n}, \mat{M} \in \mathbb{R}^{T \times T}$ denote the ALiBi matrix and the attention mask, respectively, which will be defined later.

Then, $\mat{Y} = \param{\SA_\ell(} \mat{X} \param{)}$ such that:
\begin{align}
    \mat{Q^n} &= \param{\mat{W_\ell^{n, q}}} \mat{X} + \param{\vect{b_\ell^{n, q}}} & \forall n \\
    \mat{K^n} &= \param{\mat{W_\ell^{n, k}}} \mat{X} + \param{\vect{b_\ell^{n, k}}} & \forall n \\
    \mat{V^n} &= \param{\mat{W_\ell^{n, v}}} \mat{X} + \param{\vect{b_\ell^{n, v}}} & \forall n
\end{align}
\begin{align}
   \mat{\bar{S}^n} &= \mat{A^n} + \frac{{\mat{K^n}}^\top \mat{Q^n}}{\sqrt{D^n}} & & \in \mathbb{R}^{T \times T} & \forall n \\
   \mat{S^n} &= \drop^{p_{at}}( \softmax( \mat{\bar{S}^n} \odot \mat{M} ) ) & & \in \mathbb{R}^{T \times T} & \forall n \\
   \mat{\bar{Y}^n} &= \mat{V^n} \mat{S^n} & & \in \mathbb{R}^{D^n \times T} & \forall n \\
   \mat{Y} &= \drop^{p_h}(( \sum_{n = 1}^N \param{\mat{U_\ell^n}} \mat{\bar{Y}^n}) + \param{\vect{c_\ell}} ) & & \in \mathbb{R}^{D \times T}
\end{align}
 where $\param{\mat{W_\ell^{n, q}}}, \param{\mat{W_\ell^{n, k}}}, \param{\mat{W_\ell^{n, v}}} \in \mathbb{R}^{D^n \times D}$, $\param{\mat{U_\ell^n}} \in \mathbb{R}^{D \times D^n}$, $\forall n$ are the trainable weight parameters, $\param{\vect{b_\ell^{n, q}}}, \param{\vect{b_\ell^{n, k}}}, \param{\vect{b_\ell^{n, v}}} \in \mathbb{R}^{D^n}$,  $\forall n$, $\param{\vect{c_\ell}} \in \mathbb{R}^{D}$, are the trainable bias parameters, and $p_{at}, p_h \in [0,1)$ are the \emph{at}tention and \emph{h}idden unit dropout probabilities. 

The ALiBi matrix $\mat{A^n} = [a^n_{i,j}]_{i,j} \in \mathbb{R}^{T \times T}$ is constructed as:
\begin{align}
    \tilde{N} &= 2^{\lfloor \log_2(N) \rfloor}\\
    \tilde{n} &= 1 + ((n-1) \modop \tilde{N}) - 0.5 \left\lfloor\frac{n-1}{\tilde{N}}\right\rfloor\\
    a^n_{i, j} &= 2^{- \frac{8}{{N}} \tilde{n}} \cdot (i-j) \cdot \mathbbm{1}(i < j) & \forall i,j \in [T], n \in [N]
\end{align}
and the attention mask $\mat{M} = [m^n_{i,j}]_{i,j} \in \mathbb{R}^{T \times T}$ is constructed as:
\begin{align}
    m_{i,j} &= \mathbbm{1}(i \leq j) - \infty \cdot \mathbbm{1}(i > j) & \forall i,j \in [T]
\end{align}
where we follow the convention that $\infty \cdot 0 = 0$.

\subsection{LayerNorm (\texorpdfstring{$\LN$}{LN})}

LayerNorm, $\param{\LN^\theta}: \mathbb{R}^D \rightarrow \mathbb{R}^D$, is defined as follows:
\begin{align}
    \vect{y} = \param{\LN^\theta (}\vect{x}\param{)} &= \frac{\vect{x} - \mu (\vect{x})}{\sqrt{\sigma^2(\vect{x}) + \epsilon}} \odot \param{\vect{\gamma^\theta}} + \param{\vect{\beta^\theta}}
\end{align}
where
\begin{align}
    \mu (\vect{x}) &= \frac{1}{D} \sum_i x_i &&  \in \mathbb{R} \\
    \sigma^2(\vect{x}) &= \frac{1}{D} \sum_i (x_i - \mu(\vect{x}))^2 && \in \mathbb{R} 
\end{align}
 and, $\param{\vect{\gamma^\theta}}, \param{\vect{\beta^\theta}} \in \mathbb{R}^D$ are the trainable gain and bias parameters, and $\epsilon \in \mathbb{R}$ is a small constant.

$\theta$ is used as the parametrization variable to emphasize $\param{\LN^{em}}$, $\param{\LN^f}$, and $\param{\LN^{in}_\ell}$, $\param{\LN^{at}_\ell}$, $\forall \ell$ have different (untied) $\vect{\gamma}$ and $\vect{\beta}$ parameters. 

\subsection{FeedForwardNetwork (\texorpdfstring{$\FFN$}{FFN})}

Feedforward network component $\param{\FFN_\ell}: \mathbb{R}^D \rightarrow \mathbb{R}^D$ is defined as a simple multilayer perceptron. $\vect{y} = \param{\FFN_\ell(} \vect{x} \param{)}$ such that:
\begin{align}
    \vect{h} &= \gelu( \param{\mat{W_\ell^{f}}} \vect{x} + \param{\vect{b_\ell^{f}}} ) && \in \mathbb{R}^{D'}\\
    \vect{y} &= \drop^{p_f}( \param{\mat{U_\ell^{f}}} \vect{h} + \param{\vect{c_\ell^{f}}} ) && \in \mathbb{R}^D
\end{align}
where $\gelu(x) = 0.5 \cdot x \cdot (1 + \tanh( 0.79788456 \cdot x \cdot (1 + 0.044715 \cdot x^2) ))$ is applied element-wise,
$\param{\mat{W_\ell^{f}}} \in \mathbb{R}^{D' \times D}$, $\param{\mat{U_\ell^{f}}} \in \mathbb{R}^{D \times D'}$ are the trainable weight parameters, $\param{\vect{b_\ell^{f}}} \in \mathbb{R}^{D'}$, $\param{\vect{c_\ell^{f}}} \in \mathbb{R}^D$ are the trainable bias parameters, and $p_f \in [0,1)$ denotes the dropout probability at this component.

\subsection{List of All Trainable Parameters}
\label{sec:params}

List of shape hyperparameters and their values are as follows:
\begin{itemize}
  \itemsep0 em
  \item $L = 70$ (number of layers)
  \item $N = 40$ (number of heads)
  \item $|\mathcal{V}| = 131072$ (vocabulary size)
  \item $D = 7,680$ (hidden dimension)
  \item $D^n = 192$, $\forall n \in [N]$ (hidden dimension of each head)
  \item $D' = 4D = 30,720$ (hidden dimension of $\FFN$)
\end{itemize}

\noindent Initialization hyperparameters are as follows:
\begin{itemize}
    \item $z = 0.006588 \approx 1/\sqrt{3D}$ is the default range (standard deviation).
    \item $z' = z \cdot (1/{\sqrt{2L}})$ is the rescaled range for the second layer in $\FFN$ and the final linear map in $\SA$.
\end{itemize}

\noindent List of all parameters with their sizes and (element-wise) initialization:

\noindent
\begin{tabular}{c|c|c|c|r|r|c}
Range & Group & Param & Shape & Size & Total size & Init \\
\hline
& & $\param{\mat{W^{em}}}$ & ${D \times |\mathcal{V}|}$ & 1,006,632,960 & 1,006,632,960  & $\sim \mathcal{N}(0, z)$\\
\hline
& $\param{\LN^{em}}$ & $\param{\vect{\gamma^{em}}}$ & $D$ & 7,680 & 7,680 & $=1$\\
& & $\param{\vect{\beta^{em}}}$ & $D$ & 7,680 & 7,680 & $=0$\\
\hline
$\ell \in [70]$ & $\param{\LN_\ell^{in}}$ & $\param{\vect{\gamma_\ell^{in}}}$ & $D$ & 7,680 & 537,600 & $=1$ \\
                & & $\param{\vect{\beta_\ell^{in}}}$ & $D$ & 7,680 & 537,600 & $=0$\\
\hline
$\ell \in [70]$, & $\param{\SA_\ell}$ & $\param{\mat{W_\ell^{n, q}}}$ & $D^n \times D$ & 1,474,560 & 4,128,768,000 & $\sim \mathcal{N}(0, z)$\\
 $n \in [40]$    & & $\param{\mat{W_\ell^{n, k}}}$ & $D^n \times D$ & 1,474,560 & 4,128,768,000 & $\sim \mathcal{N}(0, z)$\\
                 & & $\param{\mat{W_\ell^{n, v}}}$ & $D^n \times D$ & 1,474,560 & 4,128,768,000 & $\sim \mathcal{N}(0, z)$\\
                 & & $\param{\mat{U_\ell^n}}$      & $D \times D^n$ & 1,474,560 & 4,128,768,000 & $\sim \mathcal{N}(0, z')$\\
                 & & $\param{\vect{b_\ell^{n, q}}}$ & $D^n$ & 192 & 537,600 & $=0$\\
                 & & $\param{\vect{b_\ell^{n, k}}}$ & $D^n$ & 192 & 537,600 & $=0$\\
                 & & $\param{\vect{b_\ell^{n, v}}}$ & $D^n$ & 192 & 537,600 & $=0$\\
 \hline
$\ell \in [70]$ & $\param{\SA_\ell}$ & $\param{\vect{c_\ell}}$      & $D$ & 7,680 & 537,600 & $=0$\\
\hline
$\ell \in [70]$ & & $\param{\vect{\gamma_\ell^{at}}}$ & $D$ & 7,680 & 537,600 & $=1$\\
                & & $\param{\vect{\beta_\ell^{at}}}$ & $D$ & 7,680 & 537,600 & $=0$\\
\hline
$\ell \in [70]$ & $\param{\FFN_\ell}$ & $\param{\mat{W_\ell^{f}}}$  & $D' \times D$ & 235,929,600 & 16,515,072,000 & $\sim \mathcal{N}(0, z)$\\
                & & $\param{\mat{U_\ell^{f}}}$  & $D \times D'$ & 235,929,600 & 16,515,072,000 & $\sim \mathcal{N}(0, z')$\\
                & & $\param{\vect{b_\ell^{f}}}$ & $D'$ & 30,720 & 2,150,400 & $=0$\\
                & & $\param{\vect{c_\ell^{f}}}$ & $D$ & 7,680 & 537,600 & $=0$\\
\hline
& $\param{\LN^f}$ & $\param{\vect{\gamma^{f}}}$ & $D$ & 7,680 & 7,680 & $=1$\\
& & $\param{\vect{\beta^{f}}}$ & $D$ & 7,680 & 7,680 & $=0$\\
\hline
& & & & & 50,558,868,480 & \\
\end{tabular}

\section{Details on external financial tasks}
\label{app:external-tasks}

\begin{table}[t]
\small
\centering
\begin{tabular}{@{}p{0.2\linewidth}@{} p{0.8\linewidth}@{}}
\toprule
Tag & Question \\
\midrule
price or not & Does the news headline talk about price(?) \\
price up & Does the news headline talk about price going up(?) \\
price stable & Does the news headline talk about price staying constant(?) \\
price down & Does the news headline talk about price going down(?) \\
past price & Does the news headline talk about price in the past(?) \\
future price & Does the news headline talk about price in the future(?) \\
past general & \footnotesize{Does the news headline talk about a general event (apart from prices) in the past(?)} \\
future general & \footnotesize{Does the news headline talk about a general event (apart from prices) in the future(?)} \\
asset comparison & Does the news headline compare gold with any other asset(?) \\
\bottomrule
\end{tabular}
\caption{Official documentation of each tag~\citep{DBLP:journals/corr/abs-2009-04202}.}
\label{tab:headline-question}
\end{table}

\paragraph{FPB}~\citep{DBLP:journals/jasis/MaloSKWT14}:
The Financial Phrasebank Dataset includes a sentiment classification task on $\roughly 5,000$ sentences in the English language taken from financial news about companies listed on OMX Helsinki.
Sentiment annotations of positive, negative, neutral are adjudicated from the perspective of an investor: any news that could benefit/hurt an investor is considered positive/negative and neutral otherwise.
Each sentence is annotated by 5 to 8 annotators who have sufficient knowledge of finance, whereas the source sentences were written by financial journalists.
For example, news about shrinking revenue would be labeled negative and company growth as positive.
While there are different configurations of this dataset with each configuration denoting the percentage agreement between annotators ($\ge$$50\%$, $\ge$$66\%$, $\ge$$75\%$, $100\%$), we choose to use the configuration with $\ge$$50\%$.
Since an official train-test split is not available, we create our own random split. 
Our training split contains 3,876 sentences with 1,086 positive, 488 negative, and 2,302 neutral sentences and our test set contains 970 sentences with 277 positive, 116 negative, and 577 neutral sentences.
We choose 5 shots and report F1 score weighted by support.
\paragraph{FiQA SA}~\citep{DBLP:conf/www/MaiaHFDMZB18}:
The second sentiment analysis task is to predict the aspect-specific sentiment in English financial news and microblog headlines, which were published as a part of the 2018 challenge on financial question answering and opinion mining.
In the original task, sentiment is annotated on a continuous scale of $[-1,+1]$; the details on the annotation task are not readily available.
To make this regression dataset amenable for few-shot LLM setup, we convert it into a classification task: Negative ($-1 \le{} x < -0.1$), neutral ($-0.1 \le{} x < +0.1$), and positive ($+0.1 \le{} x \le{} +1$), where $x$ is the original sentiment score.
We selected this discretization based on a manual examination of the dataset.
Like with FPB, we create our own random split combining both microblogs and news. 
After discretization, our training set contains 938 sentences with 576 positive, 287 negative, and 75 neutral sentences and our test set contains 235 sentences with 141 positive, 76 negative, and 18 neutral sentences.
We select 5 shots and report weighted F1.

\paragraph{Headline}~\citep{DBLP:journals/corr/abs-2009-04202}:
This is a binary classification task of whether a news headline in the gold commodity domain includes certain information.
This human-annotated dataset consists of 11,412 English news headlines from 2000 to 2019 about ``gold'' scraped from providers such as Reuters, The Hindu, The Economic Times, Bloomberg, and from aggregator sites such as Kitco, and MetalsDaily.
Each news article carries a subset of the following tags: ``price or not'', ``price up'', ``price down'', ``price stable'', ``past price'', ``future price'', ``past general'', ``future general'', ``asset comparison''.
The dataset is created using annotator consensus and Cohen’s Kappa for each of the categories is $\ge$$0.85$ indicating a high-quality dataset.
Like with FPB, we create our own random split. Our training set contains 9,129 sentences with 7,780, 3,785, 3,392, 414, 7,482, 299, 1,285, 67, 1,696 examples of ``price or not'', ``price up'', ``price down'', ``price stable'', ``past price'', ``future price'', ``past general'', ``future general'', ``asset comparison'' classes, respectively.
Similarly, the test set contains 2283 sentences with 1,955, 962, 838, 109, 1,873, 82, 313, 15, 454 examples of the same classes.
We verbalize each tag into a question using the official documentation on each tag as shown in \cref{tab:headline-question}.
We used 5 shots, and report the average weighted F1 score across all categories.

\paragraph{NER}~\citep{salinas-alvarado-etal-2015-domain}:
This is a named entity recognition task on financial data gathered for credit risk assessment.
The dataset consists of 8 documents with $\roughly $55,000 words of financial agreements filed with the SEC.
The annotated entity types follow the standard CoNLL format~\citep{tjong-kim-sang-de-meulder-2003-introduction} and are annotated with PER, LOC, ORG, and MISC.
We use Fin-5 as the training data for context sampling and test on the Fin-3 split.
As MISC cannot be defined on its own but ``names (that) are not already in the other categories'' \citep{tjong-kim-sang-de-meulder-2003-introduction}, we drop all entities with type MISC.
Additionally, as it is nontrivial to learn to predict empty output in the few-shot set-up, we drop sentences that do not contain any entity.
After preprocessing, our training set contains 504 sentences with 168 PER, 745 LOC, and 241 ORG, and our test set consists of 98 sentences with 39 PER, 216 LOC, and 56 ORG.
We found that all the models required more shots to perform well. Hence, we selected 20 shots and report the entity-level F1 score.

\paragraph{ConvFinQA}~\citep{chen-etal-2022-convfinqa}:
Given an input that includes text and at least one table with financial data, the task is to answer conversational questions that require numerical reasoning over the input.
The source data is earning reports of S\&P 500 companies and consists of 3,892 conversations consisting 14,115 questions.
This task requires numerical reasoning, an understanding of structured data and financial concepts, and a model needs to relate follow-up questions to the dialog turns.
To solve this task, we use ``1 shot'' where an entire gold conversation and its context is input to the models.
In addition, as each ``turn'' of the conversation concludes, the ``turn'' along with the ``gold'' answer for that turn is appended as context for future turns.
Tables are linearized in the context (as suggested by the authors) as Markdown tables, and we replace an empty entry with "-".
The reported score is the exact match accuracy of the direct answer produced by a model. 
As test set labels are not publicly available, we report results on the dev set instead.
Our training set contains 11,104 conversations and 45,888 questions and our test set contains 1,490 conversations and 5,932 questions.

\section{Training Chronicles}
\label{sec:chron}

\setcounter{subsection}{-1}

\subsection{Still\texorpdfstring{\emojileaf}{}}

\begin{figure}[b!]
    \centering
    \includegraphics[width=\textwidth]{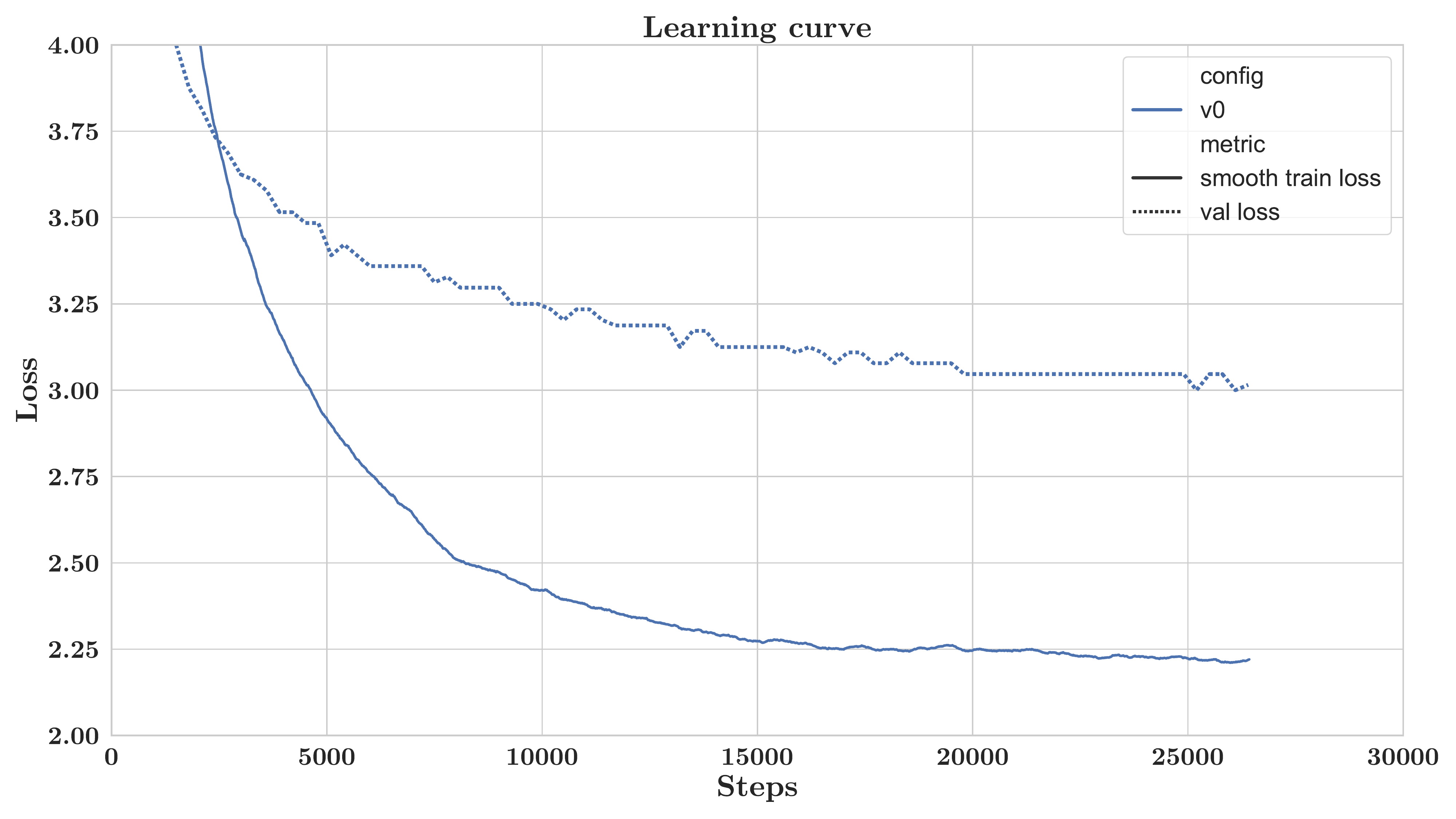}
    \caption{Learning curve of our first training attempt named v0. Observe the large gap between training and validation losses, as well as the flatness of both curves after step 20k. The final 6k steps lasted about $\roughly 2.3$ days.}
    \label{fig:chron-learning-v0}
\end{figure}

Our first training run was called v0. In this run, we experimented with curriculum learning. Data that the model would see in the future would likely be similar to the newer data in our training corpus, so we wanted the model to do better on those future documents. Additionally, since there are facts that change over time, newer information should ideally override the old. Therefore, we temporally ordered the training data by month in \dataset. 

\cref{fig:chron-learning-v0} shows the learning curve for run v0.  We observed a large gap between training and validation losses, which was expected: early stages of training would observe the oldest data (starting from 2007) whereas our validation set was strictly from the future (i.e., 2022). However, one week into training we found the model stuck on both training and validation loss, as seen by the very limited validation progress between steps 15k-20k and almost no progress after step 20k. There was the possibility that the training loss and the divergence of training and validation loss would both resolve themselves as the training data became more and more similar to the validation data as the curriculum progressed. However, we deemed this to be too risky to catch any other potential problems with the training that might require early intervention, since it would mean training for many steps without any diagnostic signal. We thus decided to abandon curriculum learning altogether.

We removed curriculum learning by shuffling all of our training data uniformly on the shard level.\footnote{Instead of loading one shard of data at a time, we load multiple random shards (without replacement) at the same time and shuffle them on the fly.} We then started a new run (v1.0), which led to much faster improvements in the validation loss. We were unable to ascertain if curriculum learning had a negative impact on training or if the loss plateaued due to other factors, for example, the other discovered issue in v1.x.

\begin{figure}[t]
    \centering
    \includegraphics[width=\textwidth]{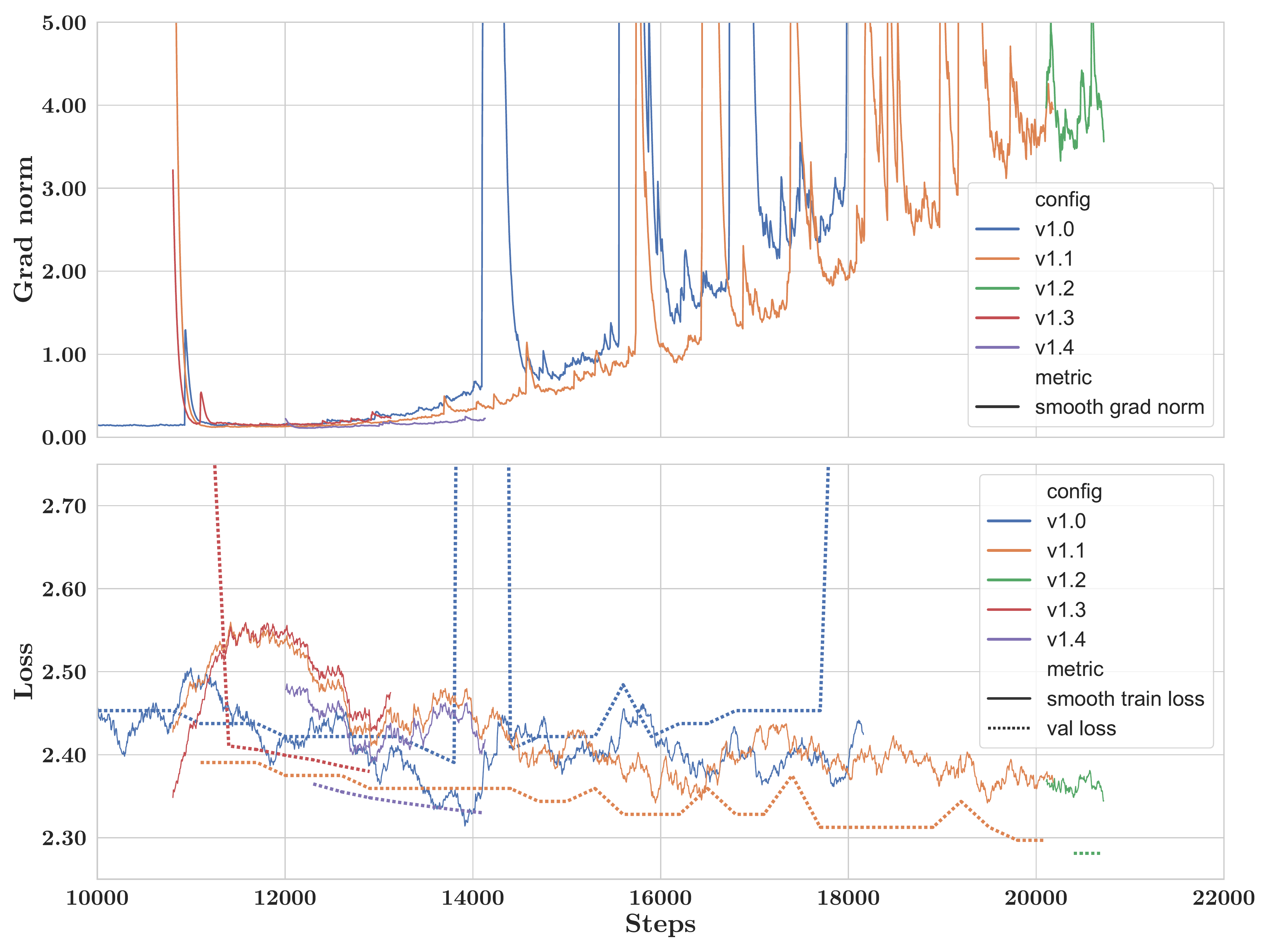}
    \caption{Gradient norms (top) and train \& validation loss (bottom) of v1.x runs.}
    \label{fig:chron-v1-grad}
\end{figure}

\subsection{Elbow\texorpdfstring{\emojielbow}{}}
\label{sec:elbow}

During our new run without curriculum learning (v1.0), we observed that the gradient norm showed a steady increase
after
about 12k steps (\roughly 4.5 days of training), with occasional spikes (see \cref{fig:chron-v1-grad}). This was accompanied by sudden jumps in the validation loss, possibly indicating that the model might be becoming sensitive to small changes in its weights. Training loss seemed to have been plateauing again, as well.

We believed that the gradient norm increases were the cause of the validation loss problems (notice the alignment between sudden validation loss jumps with some of the sudden gradient norm jumps for v1.0, in \cref{fig:chron-v1-grad}). We made several attempts across several model runs to fix the gradient norm increases:

\begin{spacing}{1}
\begin{tabular}{c|m{0.7\textwidth}}
    \toprule
    \textbf{Run} & \textbf{Changes from v1.0 run} \\
    \midrule
    Shared Change & \small - Fully shuffle any data not seen by the model checkpoint that we chose to start (or restart) from instead of shard-level shuffling\\
    \midrule
    v1.1 & \small
               - Start from step 10.8k of v1.0, prior to any gradient increase \newline
               - Reduce max learning rate from 1e-4 to 8e-5\\
    \midrule
    v1.2 & \small
                - Continue from step 20.1k of v1.1 (most recent checkpoint)\newline
                - Reduce max learning rate from 1e-4 to 6e-5\newline
                - Reduce gradient clip from 1.0 to 0.3\\
    \midrule
    v1.3 & \small
            - Start from step 10.8k of v1.0, prior to any gradient increase \newline
            - Use FP32 precision for LM-head computation (prior to softmax)\\
    \midrule
    v1.4 & \small
        - Start from step 12.0k of v1.3\newline
        - Reduce max learning rate from 1e-4 to 6e-5\newline
        - Reduce gradient clip from 1.0 to 0.3\newline
        - Use FP32 precision for LM-head computation (prior to softmax)\\
    \bottomrule
\end{tabular}
\end{spacing}

\begin{figure}[t]
    \centering
    \includegraphics[width=\textwidth]{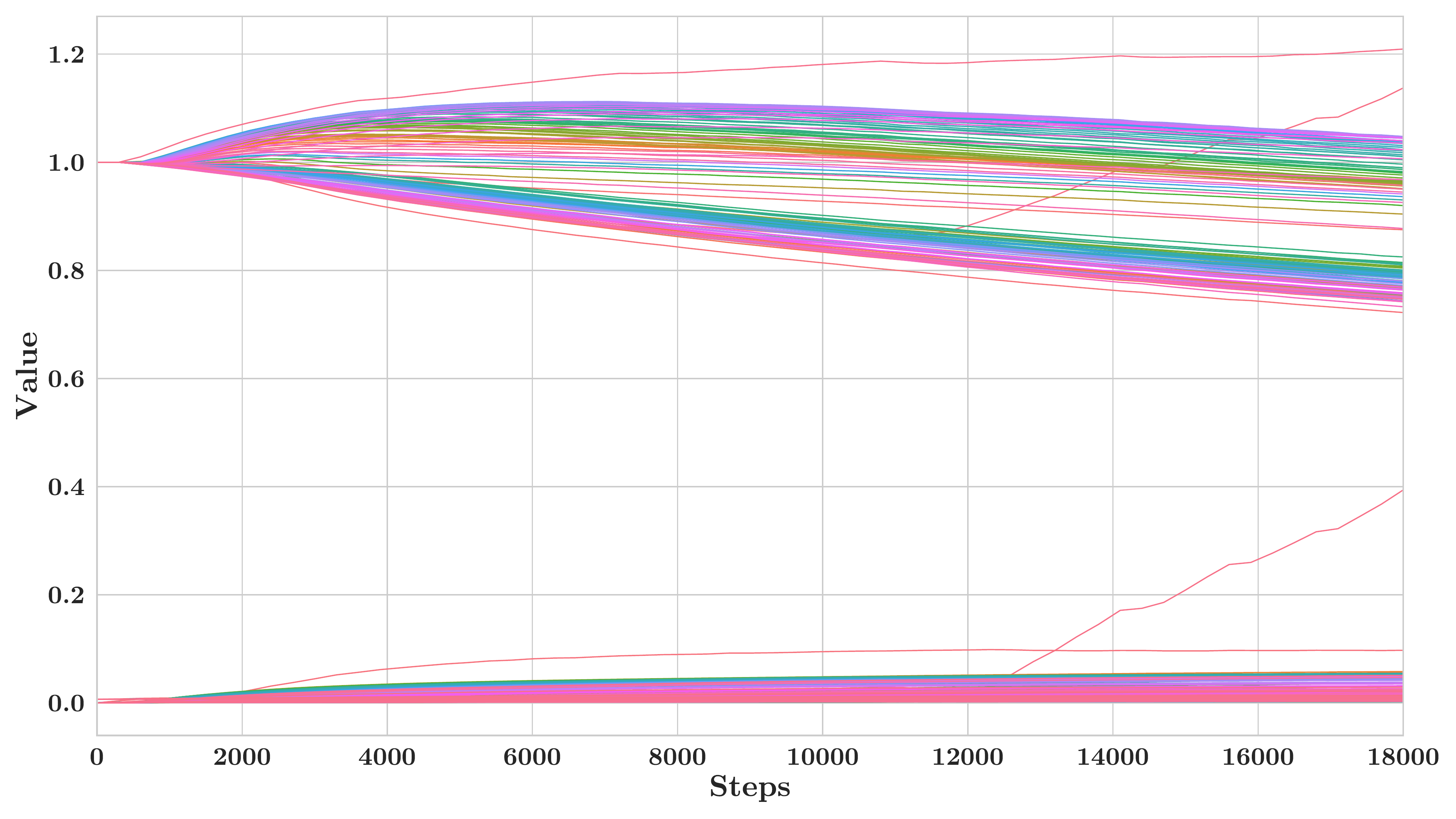}
    \caption{Rescaled norms for each component in v1.0 run. Input LayerNorm at Layer 1 stood out.}
    \label{fig:chron-v1-norm}
\end{figure}

\begin{figure}[t]
    \centering
    \includegraphics[width=\textwidth]{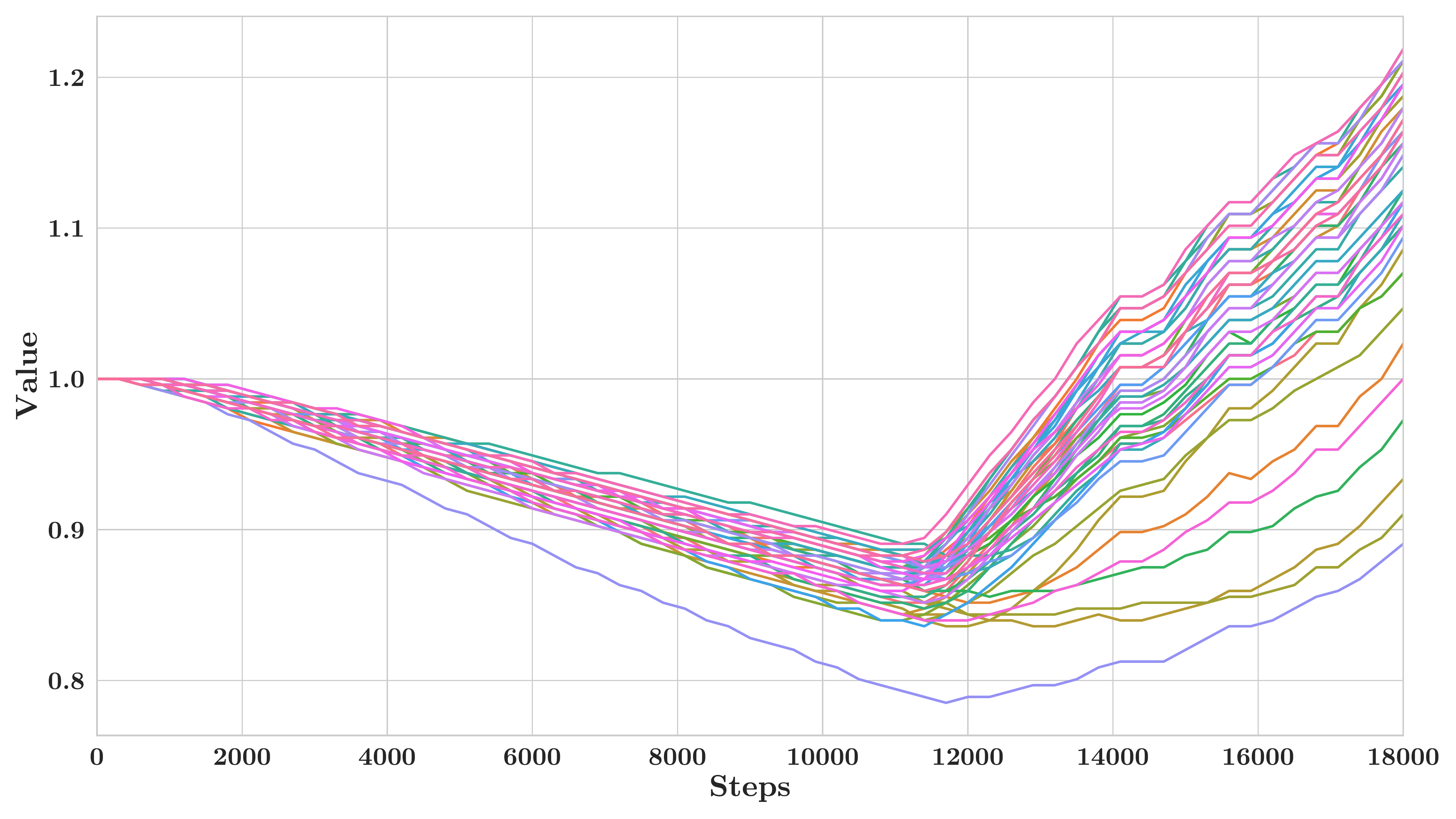}
    \caption{Values for Input LayerNorm at Layer 1 in v1.0 run.} 
    \label{fig:chron-v1-value}
\end{figure}

All of these attempted fixes were made after we observed a trend of increasing gradient norms similar to the original run (v1.0), or some early signs of a similar path that we hypothesized would eventually grow more. Since we didn't want to waste training time, we did our best to make decisions early instead of allowing the model to continue down a bad training path.

We investigated the norms of the weights themselves to see if any peculiar trends were aligning with the gradient growth. In particular, we were curious to see if there were particular layers or components that were responsible for the large gradient norms.

\cref{fig:chron-v1-norm} plots L2 norms for each component, averaged by the square root of the number of elements (layer norm multipliers start from 1 and all the others start close to zero due to initialization). We observed that all components follow a similar benign trend except one: Input LayerNorm at layer 1 (i.e. $\LN^{in}_1$), which suddenly elbows and starts increasing roughly linearly after step \roughly 12k. This also aligns with the initial growth of the gradient norms.

To take a closer look, we inspected individual values of the multiplier weights $\gamma^{in}_1$ (there are 60 such values in a single model shard out of 128) in \cref{fig:chron-v1-value}. We observed all values contributing to the same trend of shrinking until steps 11-12k and then shifting to move upward instead.

During this investigation, we discovered another bug: Weight decay was applied to all the non-bias parameters, as opposed to skipping the LayerNorm multiplier weight $\gamma$ since they are initialized at 1. To the best of our knowledge, this practice has been inherited from the BERT implementation\footnote{\url{https://github.com/google-research/bert/blob/eedf5716ce1268e56f0a50264a88cafad334ac61/optimization.py\#L59-L65}}. 

This makes the elbow artifact shown in \cref{fig:chron-v1-value} even more confusing: An additional push of weights towards 0 would trivially explain a downward trend but not a sudden shift to growth. 

After four failed attempts to fix the run, we considered the possibility of this run being unsalvageable and contemplated starting from scratch to apply a more conservative hyperparameter setting from the beginning. These would include things that we have tried in our attempts such as shrinking the learning rate or gradient clipping. Additionally, because the pathological trend change is isolated to $\LN^{in}_1$, which is topologically very close to the removed LayerNorm at the embedding layer ($\LN^{em}$) we decided to add back $\LN^{em}$ as an additional precaution, despite most other LLMs not having this component.

\subsection{Slide\texorpdfstring{\emojislide}{}}

\begin{figure}[t]
    \centering
    \includegraphics[width=\textwidth]{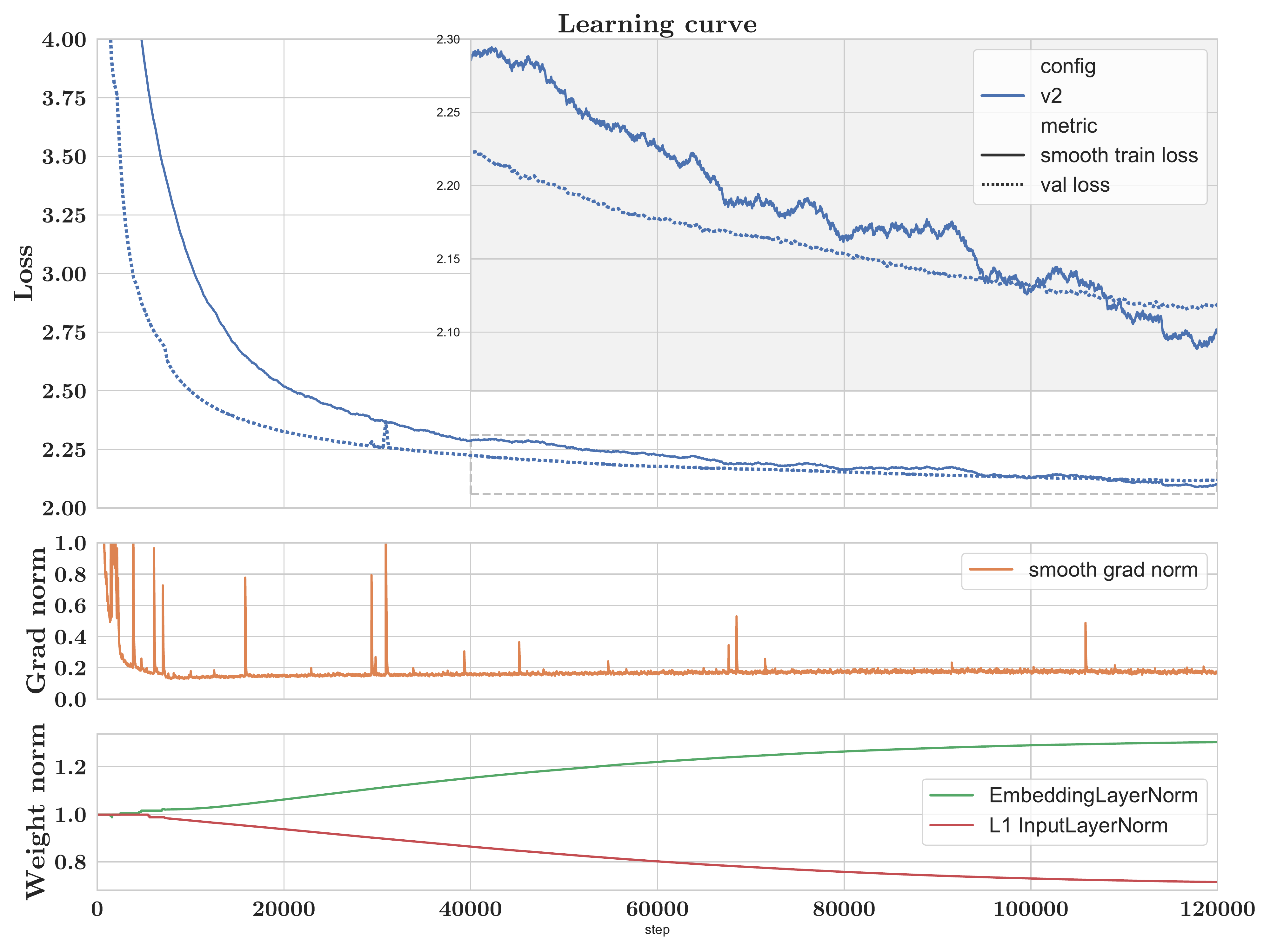}
    \caption{Loss, gradient norm and weight norms for listed components for the v2.0 run.}
    \label{fig:chron-v2-loss}
\end{figure}

After numerous attempts to fix the elbow issue, we wanted to be as conservative as possible for the hyperparameter choices when starting from scratch to keep the learning dynamics as stable as possible. We started the next run (v2.0) with the following hyperparameters and changes:

{
\begin{itemize}
    \itemsep0 em
    \item Use \full precision in LM-head ($\softmax$ in \cref{eq:lm-head})
    \item Use max learning rate of 6e-5 instead of 1e-4
    \item Use a gradient clipping value of 0.3 instead of 1.0
    \item Fully shuffle data
    \item Use a different seed to ensure different initialization and data order
    \item Reintroduce LayerNorm at embedding layer ($\LN^{em}$)
    \item Use a longer learning rate warm-up period of 1800 steps
    \item Remove incorrect use of weight decay on LayerNorm multipliers ($\gamma^{\bigcdot}_{\bigcdot}$)
    \item Use Megatron initialization rescaling (see use of $z'$ in \cref{sec:params})
    \item Apply \texttt{query\_key\_layer\_scaling} \citep{shoeybi2019megatron}
    \item Apply a batch size warm-up: Use a batch size of 1024 for 7200 iterations, then increase to 2048
\end{itemize}
}

\noindent In addition to hyperparameter changes, we also performed additional monitoring to catch issues earlier. Because we observed the pathological behavior at the first LayerNorm component, we started monitoring the norms of the weights $\gamma^{em}$ and $\gamma^{in}_1$ (scaled by $1/\sqrt{D}$).

With the aforementioned conservative choice of hyperparameters during v2.0, we observed very smooth and (thankfully!) uneventful training for approximately 42 days (\roughly 115,500 iterations). We saw few surprises both in terms of training and validation performance curves (see \cref{fig:chron-v2-loss}), as well as the norms of the gradients. The only intervention needed during this period was to restart the job after 28 days, due to the underlying platform having a hard limit on the duration of the job.

During this period, we observed a smoothly decreasing validation loss (except a few jumps earlier on) until it started to flatten around 2.116 (y-axis) at the end. Running training loss has a similar trend of overall decrease with the typical jitter and random occasional increments.

We also observed that weight norms for LayerNorm components of the initial layer were smooth and stable without any immediate or long-term trend changes (see \cref{fig:chron-v2-loss}). This presents some evidence that we were not suffering from the pathological behavior we observed in v1.x in regards to LayerNorm parameters $\gamma^{em}$.

Overall, while this set of changes led to a smooth training run in v2.0, we cannot conclude which of these changes was decisive in leading to a successful training run. We defer such investigation to future work.

\subsection{Suspense\texorpdfstring{\emojisuspense}{}}

\begin{figure}[t]
    \centering
    \includegraphics[width=\textwidth]{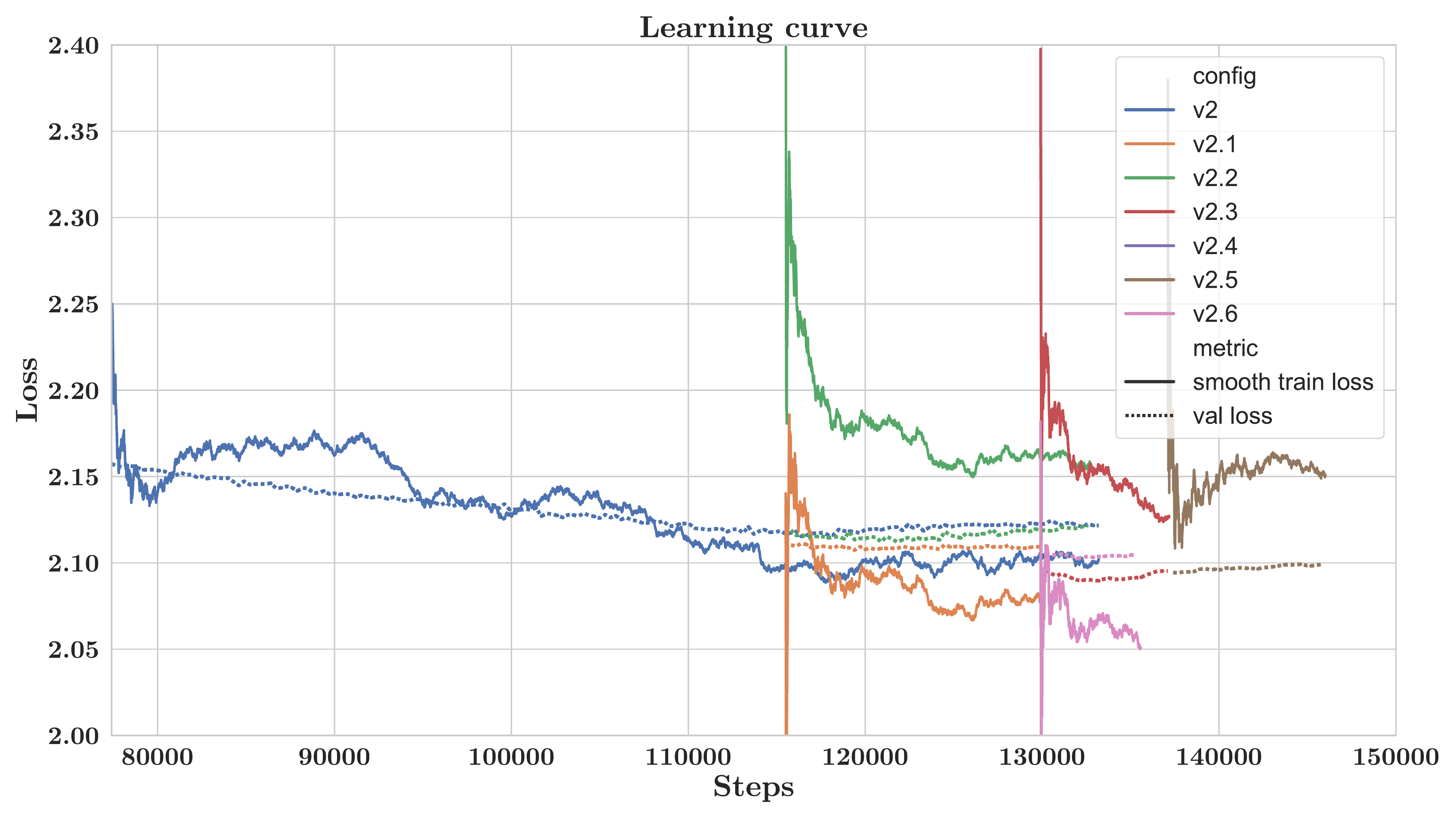}
    \caption{Loss values for various runs at the end of v2. Note v2.4 overlaps with v2.6.}
    \label{fig:chron-v3-loss}
\end{figure}

About 48 days into training v2.0, we noticed that the validation loss had not improved in a week (from iteration 115,500 to 133,200, see \cref{fig:chron-v3-loss}, v2.0 curves). During the same period, we also noticed training loss flattening around 2.10 (with the usual jitter). We suspected that the model was no longer learning properly, and decided to intervene.

We considered two options: 1) changing the max learning rate, 2) rolling back to an earlier checkpoint and re-shuffling the remainder of the data to pick up a different optimization path.

We had two proposed ways in which to change the learning rate. An argument for \emph{increasing} the learning rate was the possibility that we were stuck in a local optimum. Allowing the optimizer to make bigger jumps would allow the model to escape the optimum and continue learning. On the other hand, the argument for \emph{decreasing} the learning rate was based on \citet{opt-zhang} in which they had observed improvements after shrinking the learning rate after getting stuck. Furthermore, we had spent more steps in the high-learning-rate region of the overall learning rate schedule since, by following the Chinchilla scaling law, we had more total steps compared to models like \bloom or GPT-3.

The other option was to roll back to an earlier checkpoint, re-shuffle the remainder of the data and continue training. \citet{chowdhery2022palm} found that when they saw spikes in the validation loss, they ``re-started training from a checkpoint roughly 100 steps before the spike started, and skipped roughly 200–500 data batches.'' This suggests that data ordering mattered, and backing out of a bad data/gradient path may help. That may have been our issue with curriculum learning (v0.x), although it may have been that the issues were not with curriculum learning but with other issues we fixed in v1.0.

In the end, we decided to \emph{shrink} the learning rate, roll back to the start of the increasing validation loss trend ~7 days prior, and also re-shuffle the remaining data.

We also became concerned that our choices were based on a single, albeit large, development set. Our validation set only contained data from July 2022 (val$_\text{future}$; roughly 105M tokens), whereas the training set ranged from 2007 to June 2022, meaning that the validation set was slightly out of distribution. We had done this to ensure a future-forward evaluation, and to ensure that the training set didn't have leaked validation data. While this matched our goals, it was possible that a single month of data was not properly reflective of the model's abilities, and thus we were making decisions that overfit the validation data.
We created a second validation set from the last 105M tokens of the training set for offline evaluation (val$_\text{past}$). These tokens were from training but would be unobserved until the model finished training. However, since the model training data was fully shuffled, this validation set was not from a held-out time-period.

To assess whether a lack of progress on validation loss translates into a lack of progress on downstream evaluation performance, we used two popular benchmarks: the multiple-choice subset of BBH (bbh$_\text{sub}$) and all of MMLU. These provided additional assurance that changes in validation loss were tracking actual model improvements. Note that running a checkpoint on these benchmarks is much more time-consuming than computing the validation loss.

\begin{table}[ht!]
    \centering
    \scalebox{0.9}{
    
\begin{tabular}{r|cc|cc|cc|cc}
\toprule
  & \cellcolor[HTML]{4A86E8}{\color[HTML]{FFFFFF} \textbf{v2}} & \cellcolor[HTML]{FF9900}{\color[HTML]{FFFFFF} \textbf{v2.1}} & \cellcolor[HTML]{4A86E8}{\color[HTML]{FFFFFF} \textbf{v2}} & \cellcolor[HTML]{FF9900}{\color[HTML]{FFFFFF} \textbf{v2.1}} & \cellcolor[HTML]{4A86E8}{\color[HTML]{FFFFFF} \textbf{v2}} & \cellcolor[HTML]{FF9900}{\color[HTML]{FFFFFF} \textbf{v2.1}} & \cellcolor[HTML]{4A86E8}{\color[HTML]{FFFFFF} \textbf{v2}} & \cellcolor[HTML]{FF9900}{\color[HTML]{FFFFFF} \textbf{v2.1}} \\
\midrule
\textbf{step} & \multicolumn{2}{|c}{\textbf{val$_\text{future}$}} & \multicolumn{2}{|c}{\textbf{val$_\text{past}$}} & \multicolumn{2}{|c}{\textbf{mmlu}} & \multicolumn{2}{|c}{\textbf{bbh$_\text{sub}$}} \\
\midrule
99300 & \cellcolor[HTML]{E67C73}8.43 & & \cellcolor[HTML]{E67C73}8.60 & & \cellcolor[HTML]{FCF0F0}37.77 & & \cellcolor[HTML]{57BB8A}43.57 & \\
115500 & \cellcolor[HTML]{BFE5D2}8.30 & & \cellcolor[HTML]{FCEFEE}8.43 & & \cellcolor[HTML]{57BB8A}38.79 & & \cellcolor[HTML]{BDE4D1}43.10 & \\
126600 & \cellcolor[HTML]{FFFFFF}8.34 & \cellcolor[HTML]{57BB8A}8.24 & \cellcolor[HTML]{FFFFFF}8.40 & \cellcolor[HTML]{57BB8A}8.32 & \cellcolor[HTML]{FFFFFF}37.86 & \cellcolor[HTML]{D6EFE3}38.09 & \cellcolor[HTML]{EB9790}42.37 & \cellcolor[HTML]{FFFFFF}42.79 \\
133200 & \cellcolor[HTML]{FFFEFE}8.35 & & \cellcolor[HTML]{D0EBDE}8.38 & & \cellcolor[HTML]{E67C73}37.02 & & \cellcolor[HTML]{E67C73}42.26 &  
\\ \bottomrule
\end{tabular}

    }
    \caption{Preliminary evaluation on in-distribution (val$_\text{past}$, 105M tokens), and out-of-distribution (val$_\text{future}$; 105M tokens) validation sets (perplexity), and downstream tasks (accuracy). We report perplexity since we compare models with the same tokenization.}
    \label{tab:chron-v3-eval1}
\end{table}

Using our two dev sets and downstream evaluations for guidance, we made several attempts to improve run v2.0 and direct the model to continue learning. A summary of our attempts follows:
\begin{spacing}{1}
\centering
\begin{tabular}{c|m{0.5\textwidth}}
\toprule
\textbf{Run} & \textbf{Changes from v2.0 run} \\
\midrule
Shared Change & \small - Re-shuffle future data starting from step 115500\\
\midrule
v2.1 & \small
    - Start from v2.0 step 115500\newline
    - Reduce max learning rate from 6e-5 to 4e-5\\
\midrule
v2.2 & \small
    - Start from v2.0 step 115500\newline
    - Increase dropout from 0.0 to 0.1\\
\midrule
v2.3 & \small
    - Start from v2.1 step 129900\newline
    - Reduce max learning rate from 6e-5 to 2e-5\newline
    - Increase dropout from 0.0 to 0.1\\
\midrule
v2.4 & \small
    - Start from v2.1 step 129900\newline
    - Reduce max learning rate from 6e-5 to 2e-5\\
\midrule
v2.5 & \small
    - Start from v2.3 step 137100\newline
    - Reduce max learning rate from 6e-5 to 1e-5\newline
    - Increase dropout from 0.0 to 0.1\\
\midrule
v2.6 & \small
    - Start from v2.1 step 129900\newline
    - Reduce max learning rate from 6e-5 to 2e-5\newline
    - Reduce weight decay from 0.1 to 0.01\\
\bottomrule
\end{tabular}
\end{spacing}

After we lowered the learning rate and rolled back the model (v2.1), we observed an initial sudden (and dramatic) improvement; however, validation loss quickly flattened out. Coupled with the mixed results on downstream evaluation, we decided to enable dropout for the first time with a probability of 0.1.

With dropout, as expected, we observed a larger training loss since dropout is applied during the computation of the loss (v2.2 in \cref{fig:chron-v3-loss}). However, we observed an initially decreasing validation loss. Still, as the run progressed further, validation loss started creeping back up and met the value of the original run (v2.0, blue).

Based on these observations, we decided that further decreasing the learning rate would give us the best chance to continue learning successfully. We subsequently tried various combinations of smaller values of learning rate and adding dropout. Observe that in v2.3 (red) with 2e-5 max learning rate and in v2.5 (brown) with 1e-5 as its continuation, both with a dropout rate of 0.1, shown in \cref{fig:chron-v3-loss}. In \cref{tab:chron-v3-eval2}, we observed v2.3 led to much better perplexity and slightly better downstream performance, and v2.5 continues to improve downstream performance compared to v2.3 in the beginning, while decreasing perplexity slightly. v2.4 (purple) attempted a max learning rate of 2e-5 as well, without dropout, however. The only odd run during this time is v2.6 (pink), in which we experimented with a smaller weight decay of 0.01 (compared to the original 0.1) with a max learning rate of 2e-5 to investigate the possibility of getting stuck in local minima due to too strong of a pull from the weight decay. However, this yields almost the exact same curve as the original 0.1 weight decay (the difference between v2.4 and v2.6 is only the weight decay, and since they yield the same curve v2.6 completely hides v2.4, rendering it invisible in the plot).

In conclusion, all of the runs (summarized in \cref{fig:chron-v3-loss}) had the same outcome of eventual flattening of the validation loss and sometimes even increasing the loss.
We did not observe that any particular change significantly improved the downstream evaluations and validation loss (\cref{tab:chron-v3-eval2}).

\begin{table}[ht!]
    \centering
    \scalebox{0.75}{
    \begin{tabular}{r| cccccc|cccccc}
\toprule
 & \cellcolor[HTML]{4A86E8}{\color[HTML]{FFFFFF} \textbf{v2}} & \cellcolor[HTML]{FF9900}{\color[HTML]{FFFFFF} \textbf{v2.1}} & \cellcolor[HTML]{34A853}{\color[HTML]{FFFFFF} \textbf{v2.2}} & \cellcolor[HTML]{CC0000}{\color[HTML]{FFFFFF} \textbf{v2.3}} & \cellcolor[HTML]{9900FF}{\color[HTML]{FFFFFF} \textbf{v2.4}} & \cellcolor[HTML]{7F6000}{\color[HTML]{FFFFFF} \textbf{v2.5}} & \cellcolor[HTML]{4A86E8}{\color[HTML]{FFFFFF} \textbf{v2}} & \cellcolor[HTML]{FF9900}{\color[HTML]{FFFFFF} \textbf{v2.1}} & \cellcolor[HTML]{34A853}{\color[HTML]{FFFFFF} \textbf{v2.2}} & \cellcolor[HTML]{CC0000}{\color[HTML]{FFFFFF} \textbf{v2.3}} & \cellcolor[HTML]{9900FF}{\color[HTML]{FFFFFF} \textbf{v2.4}} & \cellcolor[HTML]{7F6000}{\color[HTML]{FFFFFF} \textbf{v2.5}} \\
\textbf{step} & \multicolumn{6}{c}{\textbf{val$_\text{future}$}} & \multicolumn{6}{|c}{\textbf{val$_\text{past}$}} \\
\midrule
99300 & \cellcolor[HTML]{E67C73}8.43 &  &  &  &  &  & \cellcolor[HTML]{E67C73}8.60 &  &  &  &  &  \\
115500 & \cellcolor[HTML]{F4C3BE}8.30 &  &  &  &  &  & \cellcolor[HTML]{F6CFCC}8.43 &  &  &  &  &  \\
126600 & \cellcolor[HTML]{F0ACA7}8.34 & \cellcolor[HTML]{FBE7E6}8.24 & \cellcolor[HTML]{F4C2BE}8.31 &  &  &  & \cellcolor[HTML]{F9DBD8}8.40 & \cellcolor[HTML]{F3FAF7}8.32 & \cellcolor[HTML]{EA9088}8.56 &  &  &  \\
131700 &  &  &  & \cellcolor[HTML]{68C296}8.09 & \cellcolor[HTML]{FFFFFF}8.20 &  &  &  &  & \cellcolor[HTML]{5ABC8C}8.22 & \cellcolor[HTML]{74C79E}8.24 &  \\
133200 & \cellcolor[HTML]{EFACA6}8.35 &  &  & \cellcolor[HTML]{57BB8A}8.08 &  &  & \cellcolor[HTML]{FBE6E4}8.38 &  &  & \cellcolor[HTML]{57BB8A}8.22 &  &  \\
137100 &  &  &  & \cellcolor[HTML]{9AD6B8}8.13 &  &  &  &  &  & \cellcolor[HTML]{D2ECDF}8.30 &  &  \\
139200 &  &  &  &  &  & \cellcolor[HTML]{ACDDC5}8.14 &  &  &  &  &  & \cellcolor[HTML]{F4FAF7}8.32\\
143400 &  &  &  &  &  & \cellcolor[HTML]{BEE4D2}8.15 &  &  &  &  &  & \cellcolor[HTML]{FFFEFE}8.33\\
145800 &  &  &  &  &  & \cellcolor[HTML]{C7E8D8}8.16 &  &  &  &  &  & \cellcolor[HTML]{FFFFFF}8.32\\
\midrule
\textbf{step} & \multicolumn{6}{c}{\textbf{mmlu}} & \multicolumn{6}{|c}{\textbf{bbh$_\text{sub}$}} \\
\midrule
99300 & \cellcolor[HTML]{F0B4AF}37.77 &  &  &  &  &  & \cellcolor[HTML]{C1E6D4}43.57 &  &  &  &  &  \\
115500 & \cellcolor[HTML]{ECF8F2}38.79 &  &  &  &  &  & \cellcolor[HTML]{FCF1F0}43.10 &  &  &  &  &  \\
126600 & \cellcolor[HTML]{F2BBB6}37.86 & \cellcolor[HTML]{F5CCC9}38.09 & \cellcolor[HTML]{F9FDFB}38.77 &  &  &  & \cellcolor[HTML]{E88B83}42.37 & \cellcolor[HTML]{F4C5C1}42.79 & \cellcolor[HTML]{F4CAC6}42.82 &  &  &  \\
131700 &  &  &  & \cellcolor[HTML]{FFFFFF}38.76 & \cellcolor[HTML]{FBECEA}38.51 &  &  &  &  & \cellcolor[HTML]{FAE5E4}43.02 & \cellcolor[HTML]{CFECDE}43.49 &  \\
133200 & \cellcolor[HTML]{E67C73}37.02 &  &  & \cellcolor[HTML]{A5DBC1}38.90 &  &  & \cellcolor[HTML]{E67C73}42.26 &  &  & \cellcolor[HTML]{D4EEE1}43.46 &  &  \\
137100 &  &  &  & \cellcolor[HTML]{FEFBFA}38.71 &  &  &  &  &  & \cellcolor[HTML]{76C8A0}44.02 &  &  \\
139200 &  &  &  &  &  & \cellcolor[HTML]{57BB8A}39.02 &  &  &  &  &  & \cellcolor[HTML]{57BB8A}44.20 \\
143400 &  &  &  &  &  & \cellcolor[HTML]{71C69D}38.98 &  &  &  &  &  & \cellcolor[HTML]{FFFFFF}43.20 \\
145800 &  &  &  &  &  & \cellcolor[HTML]{E6F5EE}38.80 &  &  &  &  &  & \cellcolor[HTML]{E3F4EC}43.37 \\
\bottomrule
\end{tabular}

    }
    \caption{Preliminary evaluation on in-distribution (val$_\text{past}$, 105M tokens), and out-of-distribution (val$_\text{future}$; 105M tokens) validation sets (perplexity), and downstream tasks (accuracy). We report perplexity here as we are comparing models with the same tokenization.}
    \label{tab:chron-v3-eval2}
\end{table}

At this point, we had used 77\% of our training data and were nearing the end of the budget we had set aside for training. Combined with all of these observations and initial promising results on the downstream benchmarks, we decided to end training despite not having gone through all of our training data. Another motivating factor was the possibility of using remaining unseen training data for subsequent runs of different styles of training and finetuning.

Based on this experience, we plan to explore different options in future experiments that have shown the potential to lead to more stable training for longer durations, including SwiGLU activations~\citep{shazeer2020glu}, RoPE embeddings~\citep{DBLP:journals/corr/abs-2104-09864}, and normalization for queries and keys in the attention layers~\citep{henry-etal-2020-query}.

\vskip 0.2in

\end{document}